\pdfoutput=1
\documentclass[11pt]{article}

\usepackage[final]{template/acl}
\usepackage{hyperref}

\usepackage{times}
\usepackage{latexsym}
\usepackage{verbatim}
\usepackage[T1]{fontenc} % For proper rendering and hyphenation of words containing Latin characters
\usepackage[utf8]{inputenc} % This assumes your files are encoded as UTF8
\usepackage{microtype}
\usepackage{inconsolata}
\usepackage{graphicx}                 
\usepackage{array}
\usepackage{booktabs}
\usepackage{multirow}
\usepackage{tabularx}
\usepackage{amssymb}% http://ctan.org/pkg/amssymb
\usepackage{pifont}% http://ctan.org/pkg/pifont
\usepackage{xcolor}
\usepackage{todonotes}
\usepackage{enumitem}
\usepackage{icomma}
\newcommand{\negation}[3]{\strut \colorbox{#3}{[#1]\textsuperscript{#2}}}
\newcommand{\affixal}[1]{\negation{#1}{affixal cue}{yellow!25}}
\newcommand{\single}[1]{\negation{#1}{single-word cue}{cyan!25}}
\newcommand{\multi}[1]{\negation{#1}{multi-word cue}{green!25}}
\newcommand{\nocue}{%
  \textsuperscript{\colorbox{gray!20}{no negation}}%
}

\usepackage{tcolorbox}

\title{Not all Negation Cues are Equal:\\
Affixal Negations Yield Better Negation Understanding}

\author{
    \textbf{Tian Tan} \and
    \textbf{Eduardo Blanco} \\
    Department of Computer Science, University of Arizona \\
    \texttt{\{tiantan, eduardoblanco\}@arizona.edu} \\
}

\begin{document}
\maketitle
\begin{abstract}

Negation remains a longstanding challenge for both  language models (LMs) and large language models (LLMs). 
Prior work mainly focuses on a small set of high-frequency single-word negation cues, such as \emph{not} and \emph{never}, with limited exploration of broader negation types and modern LLMs. 
To address this gap, we construct NegCue, a large-scale dataset containing over 1.8M samples spanning single-word, multi-word, and affixal negation with more than 200 unique cues. 
We further pre-train both encoder-only LMs and LLMs on NegCue to investigate how different negation types affect negation understanding. 
Experiments on five downstream benchmarks show that negation types contribute unevenly to performance gains under the same training scale. 
In particular, affixal negation yields the largest improvements, while the gains from the commonly studied single-word negation remain modest. 
Moreover, our results demonstrate that further pre-training improves negation understanding for both LMs and LLMs.

\end{abstract}

%%%%%%%%%%%%%%%%%%%%%%%%%%%%%%%%%%%%%%%%%%%%%%%%%%%%%%%%%%%%%%%%%%%%%%%%%%%%%%%%
%%%%%%%%%%%%%%%%%%%%%%%%%%%%%%%%%%%%%%%%%%%%%%%%%%%%%%%%%%%%%%%%%%%%%%%%%%%%%%%%
\section{Introduction}
\label{s:introduction}
%%%%%%%%%%%%%%%%%%%%%%%%%%%%%%%%%%%%%%%%%%%%%%%%%%%%%%%%%%%%%%%%%%%%%%%%%%%%%%%%
%%%%%%%%%%%%%%%%%%%%%%%%%%%%%%%%%%%%%%%%%%%%%%%%%%%%%%%%%%%%%%%%%%%%%%%%%%%%%%%%
Negation is a complex linguistic phenomenon that 
``relates an expression \emph{e} to another expression with a meaning that is in some way opposed to the meaning of \emph{e}''~\cite{sep-negation}.
Negation goes beyond mere opposites and antonyms;
nuance is often required to interpret the underlying affirmative meaning negations imply~\cite{english.grammar.2002}.
For example, ``John wasn't famous prior to his TV appearance''
implies ``John became famous because of his appearance on TV,''
which is key to answering ``How did John become famous?''

Text with negations is challenging for state-of-the-art models,
including LLMs~\cite{ye-etal-2023-assessing}.
\citet{ettinger-2020-bert} showed that BERT disregards negation cues when filling placeholders (e.g., standard language modeling),
and more recent work has shown that LLMs
%built negation-specific corpora to show that state-of-the-art systems
underperform in many tasks when negation understanding is required, including 
question answering~\cite{ravichander-etal-2022-condaqa,zhang-etal-2023-beyond}
and
information retrieval~\cite{weller-etal-2024-nevir}.

Prior proposals to improve negation understanding either
(a)~generate affirmative interpretations similar to the example above, thereby doubling the input size and requiring additional compute~\cite{hossain-blanco-2022-leveraging}
or
(b) further pre-train encoder-only LMs with a handful of selected negation cues~\cite{hosseini-etal-2021-understanding}.
Inspired by the latter,
in this paper we study the benefits of pre-training with all negation cues (affixal, single-word, multi-word)
on both encoder-only LMs as well as (smaller) LLMs.

\begin{figure}
\small
\begin{tabularx}{\columnwidth}{l X}
\toprule
\multicolumn{2}{X}{Samples from NegCue:} \\
% Affixal Negation Cue Example
% $S_1$:  & {\hspace*{1em}The room was extremely hot and crowded.}  \\
% $S_2$:  & {\hspace*{1em}Everyone felt \affixal{uncomfortable} inside.} \\
~~~$S_1^1$:  & The room was extremely hot and crowded. \\
~~~$S_2^1$:  & Everyone felt \affixal{uncomfortable} inside. \\
\addlinespace

% Single-word Negation Cue Example
% $S_1'$:  & {\hspace*{1em}The big gate was locked from the outside.}  \\
% $S_2'$:  & {\hspace*{1em}She did \single{not} manage to open it.} \\
~~~$S_1^2$:  & The big gate was locked from the outside. \\
~~~$S_2^2$:  & She did \single{not} manage to open it. \\
\addlinespace

% Multi-word Negation Cue Example
% $S_1''$:  & {\hspace*{1em}The public garden is now private property.}  \\
% $S_2''$:  & {\hspace*{1em}It is \multi{no longer} open to the public.} \\
~~~$S_1^3$:  & The public garden is now private property. \\
~~~$S_2^3$:  & It is \multi{no longer} open to the public. \\
\addlinespace

% No negation
% $S_1'''$:  & {\hspace*{1em}The weather is sunny and quite warm.} \\
% $S_2'''$:  & {\hspace*{1em}People enjoy their afternoon outside.\nocue} \\  
~~~$S_1^4$:  & The weather is sunny and quite warm. \\
~~~$S_2^4$:  & People enjoy their afternoon outside.\nocue \\  
\midrule
\end{tabularx}

Next Sentence Polarity Prediction (NSPP)

\begin{tabularx}{\columnwidth}{l XXXX}
% ===== NSPP =====
%\multicolumn{5}{l}{Next Sentence Polarity Prediction (NSPP)} \\
%\multicolumn{2}{l}{
% Input:  & $S_1$ & $S_1'$ & $S_1''$ & $S_1'''$ \\
~~~Input:  & $S_1^1$ & $S_1^2$ & $S_1^3$ & $S_1^4$ \\
~~~Output: & Yes   & Yes    & Yes     & No \\ \bottomrule
\end{tabularx}

% \multicolumn{2}{X}{Positive label: Yes, $S_2$ has negation.} \\
% \multicolumn{2}{X}{Negative label: No, $S_2$ has no negation.} \\

\caption{
  Four examples from NegCue, the corpus we create for Next Sentence Polarity Prediction,
  our self-supervised task to improve negation understanding.
  Given a sentence ($S_1^i$),
  the task is to predict whether the next sentence ($S_2^i$) contains negation.
  While previous work is limited to a handful of cues (e.g., \emph{not}, \emph{never}),
  we work with all negation types (affixal, single- and multi-word) and over 200 negation cues.
}
\label{f:introexample}
\end{figure}

Specifically, we pre-train for Next Sentence Polarity Prediction (NSPP), a recently proposed self-supervised task for negation understanding, which predicts whether a sentence is followed by a sentence containing negation.
To this end, we create NegCue, a corpus of over 1.8M samples with an equal number of affixal, single-word, and multi-word cues.
Figure \ref{f:introexample} illustrates the negation cue types, the NegCue corpus, and the NSPP task.
In this paper, we evaluate NSPP and NegCue on BERT, RoBERTa, and smaller variants of Llama and Qwen. %, although both are applicable to a broader range of models.
Our contributions are as follows:\footnote{Code and dataset available at \url{https://github.com/TT159/NegCue.git}}

\begin{enumerate}[noitemsep,topsep=0pt,parsep=0pt,partopsep=0.3pt]
  \item NegCue, a large dataset for Next Sentence Polarity Prediction (1.8M samples and 214 unique cues).
    It considers all negation cues: affixal, single-word, and multi-word.
  \item Experiments with encoder-only LMs (BERT and RoBERTa) and LLMs ranging from 0.5B to 3B parameters (Llama and Qwen) show that pre-training with NegCue improves performance across five benchmarks.
  \item Analysis showing which negation cues are most beneficial to pre-train with and which are most beneficial during inference. %and which benefits the most at inference time.
  
\end{enumerate}

%%%%%%%%%%%%%%%%%%%%%%%%%%%%%%%%%%%%%%%%%%%%%%%%%%%%%%%%%%%%%%%%%%%%%%%%%%%%%%%%
%%%%%%%%%%%%%%%%%%%%%%%%%%%%%%%%%%%%%%%%%%%%%%%%%%%%%%%%%%%%%%%%%%%%%%%%%%%%%%%%
\section{Related Work}
\label{s:related_work}
%%%%%%%%%%%%%%%%%%%%%%%%%%%%%%%%%%%%%%%%%%%%%%%%%%%%%%%%%%%%%%%%%%%%%%%%%%%%%%%%
%%%%%%%%%%%%%%%%%%%%%%%%%%%%%%%%%%%%%%%%%%%%%%%%%%%%%%%%%%%%%%%%%%%%%%%%%%%%%%%%
Negation is ubiquitous in English:
over 28\% of sentences from books and
27\% from oral and written conversations contain negation~\cite{hossain-etal-2020-analysis}.
The same is true in over 13\% of sentences from biomedical papers \cite{szarvas-etal-2008-bioscope}.
Beyond frequent single-word cues (e.g., \emph{not}, \emph{never}),
negation can be expressed in several ways,
including
affixal cues (e.g., careless, unadjusted),
infrequent single-word cues (e.g., forbid, deny),
and
multi-word cues (e.g., no longer, be deprived of).

Despite being frequent, there is plenty of evidence that negation is not quite acquired by models
during standard pre-training (language modeling, next sentence prediction, etc.).
For example, previous work has shown that
BERT mostly disregards negation~\cite{ettinger-2020-bert},
% and natural language understanding (NLU) is more challenging when negation is present in the input with 
and natural language understanding (NLU) becomes more challenging in the presence of negation for
both encoder-only LMs~\cite{singh2023nlms}
and LLMs~\cite{kim2025semantic,varshney2025investigating}.
Similarly, machine translation is more challenging if the source sentence contains negation~\cite{hossain-etal-2020-non},
as are other tasks we study: 
question answering~\cite{ravichander-etal-2022-condaqa,zhang-etal-2023-beyond}   
and 
information retrieval~\cite{weller-etal-2024-nevir}.

Previous work on making models better at processing negation---regardless of the benchmarks used for evaluation---%
targets a handful of negation cues: \emph{not}, \emph{n't}, and \emph{never}.
They can be broadly divided into two categories.
The first category includes approaches to generate underlying affirmative interpretations from negated statements
and concatenating these interpretations to the original negated sentences~\cite{hossain-blanco-2022-leveraging,rezaei-blanco-2024-paraphrasing}.
For example, given ``Extinct volcanoes have not erupted in recent history,''
they first generate the affirmative interpretation ``Extinct volcanoes erupted a long time ago''
and then feed to the model both sentences (as opposed to only the original statement with negation).
Note that these approaches increase compute at inference time.
The second category explores self-supervised tasks and pre-training with encoder-only LMs~\cite{hosseini-etal-2021-understanding,truong-etal-2022-improving}, which introduce no extra overhead during inference.

~\citet{rezaei-blanco-2025-making} proposed Next Sentence Polarity Prediction (NSPP), a simple yet effective self-supervised task for negation understanding in encoder-only LMs.
The task uses only the preceding sentence as input, and the prediction relies on the semantic and logical relationships between coherent adjacent sentences in discourse.
Prior studies on discourse coherence have shown that neighboring sentences often share semantic, logical, and emotional continuity~\cite{jurafsky2014speech,carter2024discourse,lu2025principle}. 
As a result, the preceding sentence often establishes semantic cues about whether the following sentence is likely to express affirmative or negated content, and even how such negation may be expressed.
% the preceding sentence frequently provides cues about the likely structure and polarity of the following sentence, including whether it expresses affirmative or negated content.

In this paper, we build upon this line of work.
Specifically, we explore for the first time the benefits of working with all negation cue types (affixal, single-, and multi-word).
Further, we demonstrate across five benchmarks spanning question answering, information retrieval, and natural language inference (NLI) that pre-training with NegCue benefits both LMs and LLMs. 
We also show that negation cue types are not equally beneficial; surprisingly, the commonly studied single-word negation type generally yields the smallest gains.
% The benchmarks we work with were built to assess the ability of models to understand negation.

%%%%%%%%%%%%%%%%%%%%%%%%%%%%%%%%%%%%%%%%%%%%%%%%%%%%%%%%%%%%%%%%%%%%%%%%%%%%%%%%
%%%%%%%%%%%%%%%%%%%%%%%%%%%%%%%%%%%%%%%%%%%%%%%%%%%%%%%%%%%%%%%%%%%%%%%%%%%%%%%%
\section{Improving Negation Understanding} % through Cue Type Variation}
\label{s:datasets_framework}
%%%%%%%%%%%%%%%%%%%%%%%%%%%%%%%%%%%%%%%%%%%%%%%%%%%%%%%%%%%%%%%%%%%%%%%%%%%%%%%%
%%%%%%%%%%%%%%%%%%%%%%%%%%%%%%%%%%%%%%%%%%%%%%%%%%%%%%%%%%%%%%%%%%%%%%%%%%%%%%%%
We improve negation understanding by further pre-training off-the-shelf models 
with NSPP, a self-supervised task.
We first introduce the NSPP task, and then present NegCue, the corpus used to pre-train.
NegCue contains samples of three negation types (1/3 each: affixal, single- and multi-word)
and over 200 negation cues.

\subsection{Next Sentence Polarity Prediction (NSPP)}
We adopt NSPP as our pre-training task because prior work has shown that it effectively improves negation understanding in encoder-only LMs, 
while also representing a recent self-supervised objective designed for negation understanding.
Given a sentence $S_1$ from some document,
the task consists of predicting whether the next sentence $S_2$ contains negation (i.e., the polarity of the next sentence).
As illustrated in Figure~\ref{f:introexample}, the model outputs a binary label indicating whether the next sentence has a negation cue. 
In this work, we further investigate its effectiveness not only for traditional LMs, but also for LLMs under different negation types.

As discussed in Section~\ref{s:related_work}, coherent discourse often exhibits semantic and emotional continuity between adjacent sentences. 
Consequently, the context established by one sentence may provide predictive cues about whether the following sentence contains negation. 
Such cues need not involve explicit negation in the preceding sentence, but may instead arise from its broader semantic content and discourse context. 
The preceding context may also contain information relevant to the type of negation expressed in the following sentence. 
% Thus, these contextual regularities provide a learnable training signal for NSPP.
Thus, these contextual regularities provide a learnable training signal for NSPP to model the occurrence and type of negation in subsequent sentences.

To validate this intuition, we conduct a linear probing experiment using a simple logistic regression classifier trained on only approximately 6K samples. 
Despite its simplicity, the model is able to predict whether the following sentence $S_2$ contains negation using only $S_1$ as input, %from $S_1$ alone 
while also capturing patterns associated with different negation types and achieving performance above random chance. 
Since LMs and LLMs are substantially more powerful than linear models, they should be expected to capture these signals more effectively and therefore benefit from NSPP pre-training. 
Additional details are provided in Appendix~\ref{app:nspp_validate}.
% about the linear probing experiment 

\begin{figure}
\centering
\includegraphics[width=\linewidth]{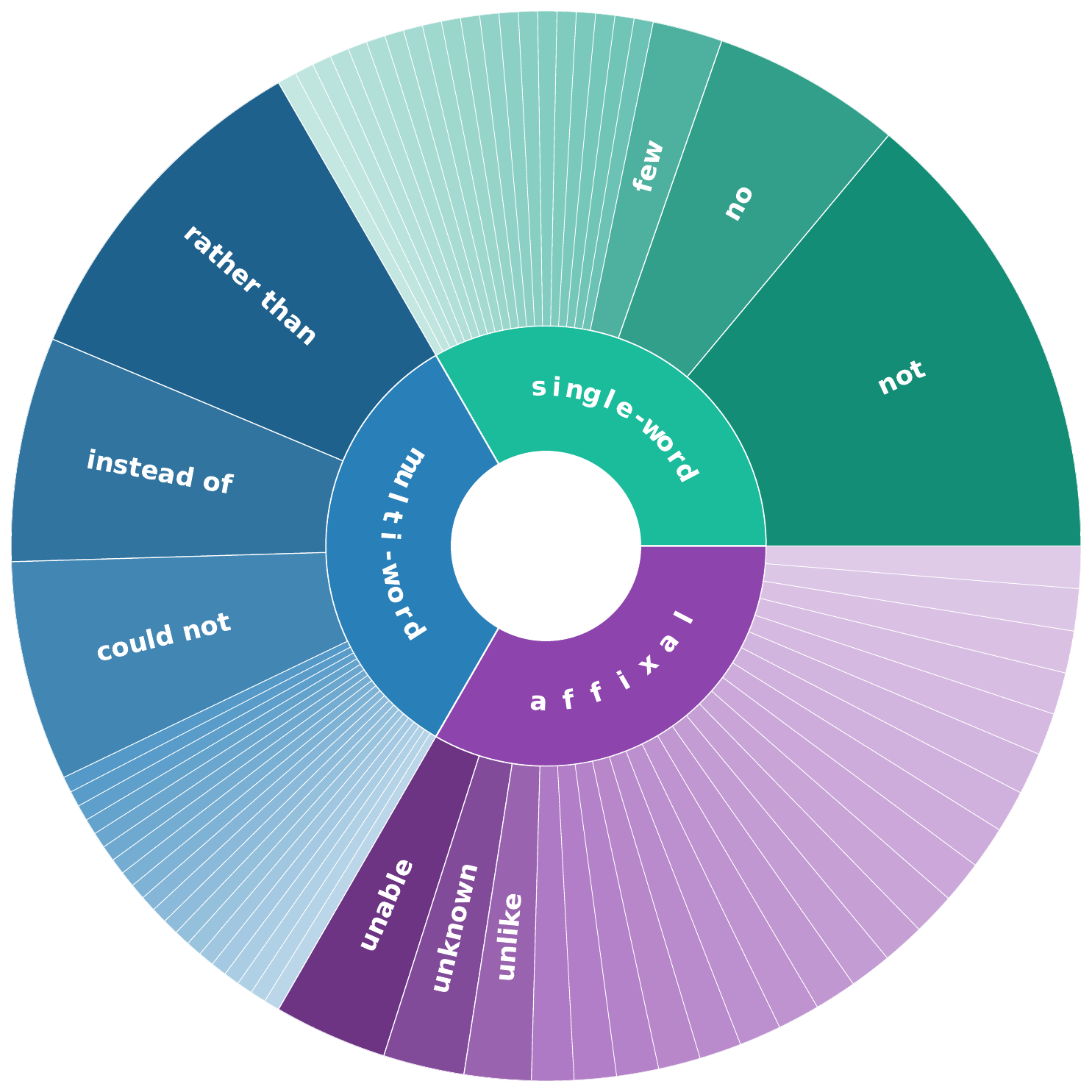}
\caption{
  Frequency distribution of the negation cues in the NegCue training splits.
  Each cue type (inner ring) accounts for 1/3 of NegCue.
  The three most frequent cues per type (outer ring) account for
  71\% and 65\% of multi- and single-word cues,
  but only 24\% of affixal cues.
  Unlabeled segments indicate the remaining cues.
  % The unlabeled segments illustrate the remaining cues.
}
\label{fig.nspp_cue_donut}
\end{figure}

\subsection{NegCue: A Large-Scale Negation Corpus}
\label{subs:negcue}
Existing negation-related corpora and tasks often rely on a very limited set of high-frequency negation cues (Section~\ref{s:related_work}). 
For example, \citet{she-etal-2023-scone} construct a dataset focusing primarily on the cue \emph{not} to study LLMs' understanding of negation scope. 
Similarly, \citet{rezaei-blanco-2025-making} consider only three negation cues: \emph{not}, \emph{n't}, and \emph{never}, to improve model robustness to negation.
 
Negation, however, is expressed in many complex and varied ways in language~\cite{horn1989natural}, appearing in
affixal (e.g., \emph{impossible}, \emph{careless}),
single-word (e.g., \emph{without}, \emph{few}),
and
multi-word (e.g., \emph{with the exception of}, \emph{no longer}) forms. 
To examine how these cue types influence negation understanding,
we construct NegCue, a corpus including consecutive sentences ($S_1$, $S_2$) from genuine documents,
where $S_2$ may or may not contain an affixal, single- or multi-word negation cue.

\paragraph{Selecting Text Sources and Negation Cues.}
The construction of NegCue relies on two key decisions:
which texts to consider and the criteria to identify negation cues.
Regarding text sources, we chose to work with a snapshot of the 2023 English Wikipedia~\cite{wikidump},
as it contains polished sentences.
Further, previous work has estimated that $\approx$9\% of sentences in Wikipedia contain negation~\cite{hossain-etal-2020-analysis}.

While understanding how negation changes meaning in natural language poses substantial challenges, 
identifying negation cues in text is much easier.
Indeed, traditional machine learning with lexical features obtains 91 F1~\cite{lapponi-etal-2012-uio}.
Inspired by this insight and previous work relying on word matching~\cite{morante-daelemans-2012-conandoyle},
we identify negation cues via keyword matching.
Specifically, we use the negation cues in CondaQA~\cite{ravichander-etal-2022-condaqa}
and the affixal cues corpus proposed by~\citet{van-son-etal-2016-building}, following the definition of~\citet{joshi2012affixal}.
Combining these corpora, we obtain 214 unique negation cues.

\paragraph{Categorizing Negation Types.}
We must know the types of each negation cue in NegCue in order to analyze the impact of each type in improving negation understanding with NSPP.
CondaQA contains affixal, single-, and multi-word negation cues,
but it does not explicitly indicate cue types.
Categorizing cues into types appears straightforward, but in practice it is non-trivial~\cite{morante2011annotation, van-son-etal-2016-building, truong-etal-2024-revisiting}. %, blanco2011some}.
For example, some single-word negations include a prefix that sometimes---but  not always---indicates an affixal negation.
Consider \emph{absence}, a single-word negation cue.
It starts with \emph{a-}, a prefix that often signals affixal cues such as \emph{atypical}.
To make things more complicated, many English words begin with \emph{a-} but are not negation cues (e.g., \emph{attention}).

We devise a deterministic approach to categorize the negation cues in NegCue.
First, cues are categorized as affixal if they are in the corpus created by~\citet{van-son-etal-2016-building}.
Second, the remaining cues are categorized as single- or multi-word based on their token counts.
Lastly, we adopt a strategy prioritizing longest matches.
This allows us to identify \emph{no longer} as a multi-word negation cue in 
``John no longer enjoys city life'' as opposed to the single-word cue \emph{no}.
Finally, we manually review the resulting classifications for quality control.

\paragraph{Obtaining Sentence Pairs and samples for NSPP.}
In order to obtain ($S_1$, $S_2$) sentence pairs,
we iterate over each pair of consecutive sentences in each Wikipedia article.
We select those such that $S_2$
(a)~contains exactly one negation cue from our list
and
(b)~is not a question.
Then, we create a negation sample ($S_1$, yes\_neg) for NSPP.
In addition, we create an affirmative sample ($S_1$, no\_neg) for NSPP
by randomly selecting from the same article a sentence followed by a sentence without negation.
If an article does not contain enough samples to balance the two classes, we collect additional samples from other articles.
This setup allows us to construct samples for each negation type, while naturally preserving the cue frequency distribution observed in real-world text. 
To enable a controlled comparison of different negation types during pre-training, we further balance the overall number of samples across the three types.

The final version of NegCue contains the same amount of samples derived from each cue type~(affixal, single-, and multi-word).
The frequency of each negation cue, however, follows the natural distribution in Wikipedia~(i.e., more frequent cues in Wikipedia appear more often in NegCue).
We also conduct a manual quality inspection of the dataset. 
Additional details are provided in Appendix~\ref{app:negcue_details}.

\begin{table}
\centering
\small
% # of NEG samples in each corpus: 278871

% \begin{tabular}{l rrr}
% \toprule
%             & \textbf{Affixal} & \textbf{Single} & \textbf{Multi} \\ 
% \midrule
% Avg. sent. length 
%             & 22.3     & 22.2       & 22.9       \\ \midrule

% \# negation cues  
%             & 167      & 36          & 11         \\

% ~~~most frequent
%             & unable   & not         & rather than \\
% ~~~~~~frequency (\%)
%             & 10.29    & 41.81       & 31.05       \\
% ~~~least frequent
%             & uncarved & forbid      & be deprived of \\
% % ~~~~~~frequency (\#)
% %             & 12     & 90       & 126  \\ \bottomrule
% ~~~~~~frequency (\%)
%             & 0.004    & 0.032        & 0.045  \\ \bottomrule
% \end{tabular}

\begin{tabular}{l ccc}
\toprule
& \multicolumn{1}{c}{\textbf{Affixal}}
& \multicolumn{1}{c}{\textbf{Single}}
& \multicolumn{1}{c}{\textbf{Multi}} \\
\midrule

Avg. sent. length
& 22.3
& 22.2
& 22.9 \\
\midrule

\# negation cues
& 167
& 36
& 11 \\

~~~most frequent
& unable
& not
& rather than \\

~~~~~~frequency (\%)
& 10.29
& 41.81
& 31.05 \\

~~~least frequent
& uncarved
& forbid
& be deprived of \\

~~~~~~frequency (\%)
& 0.004
& 0.032
& 0.045 \\
\bottomrule
\end{tabular}
\caption{
  Descriptive statistics of the NegCue training splits.
  There are more affixal cues, but most are infrequent.
  On the other hand, there are fewer single- and multi-word cues,
  but they are relatively frequent (\emph{not}: 41.81\%, \emph{rather than}: 31.05\%).
  Sentence length (\# words) is roughly the same across cue types.
}
\label{t:nspp_corpora}
\end{table}

\paragraph{Negation Cues in NegCue.}
Figure \ref{fig.nspp_cue_donut} and Table \ref{t:nspp_corpora}
analyze the negation cues in NegCue (more specifically, the cues present in $S_2$ in the negation samples).
The three cue types have the same count (33\%),
but we observe substantially different cue frequencies per type.
There are only 11 multi-word cues, and the top-3 most frequent account for 71\%.
On the other hand,
there are 167 affixal cues, and the most frequent (\emph{unable}) accounts for 10.29\%.
Single-word cues are relatively infrequent except \emph{not} (41.81\%);
similar to the affixal cues, there is a long-tail of infrequent single-word cues.

It is worth noting that sentence length is remarkably constant (22.2--22.9 words)
regardless of which cue type is present~(Table \ref{t:nspp_corpora}).
Overall, NegCue contains in the train split over 1,673,226 samples (50/50 negation/affirmative),
with 557,742 belonging to each negation cue type.
The validation and test splits contain
92,952 and 92,964 samples~(also balanced across labels and cue types).

%%%%%%%%%%%%%%%%%%%%%%%%%%%%%%%%%%%%%%%%%%%%%%%%%%%%%%%%%%%%%%%%%%%%%%%%%%%%%%%%
%%%%%%%%%%%%%%%%%%%%%%%%%%%%%%%%%%%%%%%%%%%%%%%%%%%%%%%%%%%%%%%%%%%%%%%%%%%%%%%%
\section{Experiments}
\label{s:experiments}
%%%%%%%%%%%%%%%%%%%%%%%%%%%%%%%%%%%%%%%%%%%%%%%%%%%%%%%%%%%%%%%%%%%%%%%%%%%%%%%%
%%%%%%%%%%%%%%%%%%%%%%%%%%%%%%%%%%%%%%%%%%%%%%%%%%%%%%%%%%%%%%%%%%%%%%%%%%%%%%%%

We experiment with 
encoder-only LMs (BERT-large and RoBERTa-large) 
and 
smaller LLMs, including Llama3.2-1B, Llama3.2-3B, and Qwen2-0.5B, 
and 
evaluate them on five negation-specific benchmarks. 
Additional results for the Qwen model and experimental details are provided in Appendix~\ref{app:qwen_results}. 
Specifically, we evaluate both off-the-shelf models and models further pre-trained with NSPP and NegCue. 
Section~\ref{s:pretrain-negcue} describes the further pre-training setup, 
while Section~\ref{s:evaluate-benchmarks} presents the downstream evaluation.

\subsection{NSPP Pre-training with NegCue}
\label{s:pretrain-negcue}
We further pre-train off-the-shelf versions of the LMs and LLMs with NegCue.
Specifically, we pre-train for NSPP with the full NegCue corpus
as well as subsets consisting of negation samples belonging to each negation cue type (affixal, single- and multi-word).
This setup allows us to examine how pre-training with different negation cue types affects negation understanding.

All encoder-only LMs are further pre-trained under the same settings. 
LLMs are further pre-trained using QLoRA with 4-bit quantization~\cite{dettmers2023qlora}. 
We perform minimal hyperparameter tuning and primarily use commonly adopted settings from prior practice. 
All NegCue samples are converted into instruction-style prompts for zero-shot downstream evaluation. 
Additional implementation details and the evaluation prompts are provided in Appendices~\ref{app:nspp_experimental_details} and~\ref{app:evaluation_details}, respectively.

\subsection{Evaluation on Downstream Tasks}
\label{s:evaluate-benchmarks}
We evaluate all models (off-the-shelf and further pretrained with NegCue)
in downstream tasks under two settings.
Encoder-only LMs are evaluated after supervised fine-tuning with relevant data,
and LLMs are evaluated with zero-shot prompting.

\textbf{CondaQA}
~\cite{ravichander-etal-2022-condaqa} is a large contrastive question answering benchmark (see Appendix~\ref{app:benchmarks} for additional details).
Following the original setup, we report accuracy and group consistency on the full test split. 
Group consistency measures whether all examples in a contrastive group are predicted correctly. 
We also partition the test samples into three subsets depending on the negation cue type (affixal, single-word, and multi-word)
to investigate which cue types benefit the most (at inference time) depending on which cue types are used for pre-training.
%accuracy separately for each subset.
Additionally, we partition the test samples into four quartiles based on negation cue frequency
to investigate whether the benefits of pre-training for NSPP with NegCue
apply to all negations regardless of their frequency.

For LMs, we fine-tune both off-the-shelf and further pre-trained models on the CondaQA training split.
Hyperparameters are selected via grid search over the same search space for all models, with the best configuration chosen based on validation performance.
For LLMs, we evaluate on the CondaQA test set using zero-shot in-context learning~\cite{brown2020language}.
%, where all samples are presented as instruction-style prompts.
See Appendix~\ref{app:condaqa_details} for details.

\textbf{NeQA}
~\cite{zhang-etal-2023-beyond} is a multiple-choice question answering benchmark designed for evaluating LLMs and does not provide an official training-validation split. 
To enable fine-tuning for encoder-only LMs, we construct training and validation splits based on the dataset metadata. 
Each multiple-choice question is converted into two binary classification samples by pairing the question with each answer option and assigning a label indicating whether the option is correct. 
For LLMs, we evaluate performance on the test set using zero-shot prompting and report per-example accuracy.

\textbf{NevIR}
~\cite{weller-etal-2024-nevir} is an information retrieval benchmark in which each instance consists of two queries and two documents. 
% The task evaluates whether a retrieval model assigns a higher similarity score to the relevant document for each query. 
The task evaluates whether a retrieval model assigns a higher similarity score to the relevant document for each query, rather than directly evaluating encoder-only LMs or LLMs.
%To better adapt the benchmark for encoder-only LMs and LLMs, 
To better adapt the task to our evaluation setting,
we reformulate it as a question answering task: 
each sample contains one query and two candidate documents, and the model selects the relevant document. 
We report per-example accuracy under this setting.
For LMs, we fine-tune models using the Sentence-Transformers framework~\cite{reimers-gurevych-2019-sentence} on the training set and evaluate performance on the test set. 
LLMs are evaluated using zero-shot prompting. 
% see Appendix~\ref{app:nevir_details} for the exact prompt.

\textbf{NMoNLI and ScoNe-NLI}
~\cite{geiger-etal-2020-neural,she-etal-2023-scone} are negation-focused NLI benchmarks derived from SNLI~\cite{bowman-etal-2015-large}. 
LLMs are evaluated using zero-shot prompting. 
Fine-tuning encoder-only LMs directly on their training sets leads to near-perfect performance, with test accuracies exceeding 99\% across multiple test subsets (Table~\ref{t:finetune_on_scone}), making meaningful comparisons difficult. 
We therefore fine-tune encoder-only LMs on the SNLI training set and evaluate them on the NMoNLI and ScoNe-NLI test splits.

%%%%%%%%%%%%%%%%%%%%%%%%%%%%%%%%%%%%%%%%%%%%%%%%%%%%%%%%%%%%%%%%%%%%%%%%%%%%%%%%
%%%%%%%%%%%%%%%%%%%%%%%%%%%%%%%%%%%%%%%%%%%%%%%%%%%%%%%%%%%%%%%%%%%%%%%%%%%%%%%%
\section{Results and Discussion}
\label{s:results}
%%%%%%%%%%%%%%%%%%%%%%%%%%%%%%%%%%%%%%%%%%%%%%%%%%%%%%%%%%%%%%%%%%%%%%%%%%%%%%%%
%%%%%%%%%%%%%%%%%%%%%%%%%%%%%%%%%%%%%%%%%%%%%%%%%%%%%%%%%%%%%%%%%%%%%%%%%%%%%%%%

\begin{table*}[t!]
  \centering
  \small
  \begin{tabular}{l r r@{ }l @{\ }rrr rrrr}
\toprule
&  &  \multicolumn{5}{c}{\textbf{Accuracy}}  & \multicolumn{4}{c}{\textbf{Group Consistency}} \\
\cmidrule(lr){3-7} \cmidrule(lr){8-11}
& \# Pars. & All &($\Delta$\%) & Affixal & Single & Multi  & All & Par. & Sco. & Aff.  \\ \midrule

Fine-tuned encoder-only LMs \\
~~~Previous Work \\ %~\citet{ravichander-etal-2022-condaqa}  \\
~~~~~~BERT-large [1]          & 340M & 46.3 &  & n/a & n/a & n/a & 2.2  & 44.2 & 14.8 & 12.4 \\
~~~~~~RoBERTa-large [1]       & 355M & 54.1 &  & n/a & n/a & n/a & 13.6 & 51.6 & 26.5 & 27.2 \\
~~~~~~~~~+ affirmative interpretations [2] & & 67.1 &  & n/a & n/a & n/a & 31.4 & 61.9 & 43.8 & 50.7 \\
~~~~~~~~~+ NSPP (not, n't, never) [3] & & 67.3 &  & n/a & n/a & n/a & 33.5 & 64.6 & 46.3 & 50.2 \\

~~~~~~UnifiedQA-v2-large [1]  & 770M & 66.7 &  & n/a & n/a & n/a & 30.2 & 64.0 & 43.7 & 46.5 \\
%~~~~~~UnifiedQA-v2-3B      & 3B   & 73.3 &  & n/a & n/a & n/a & 42.2 & 72.8 & 55.7 & 57.2 \\

%~~~From~\citet{rezaei-blanco-2024-paraphrasing} \\
%~~~~~~RoBERTa-large + Aff. & 355M & 67.1 &  & n/a & n/a & n/a & 31.4 & 61.9 & 43.8 & 50.7 \\
%~~~From~\citet{rezaei-blanco-2025-making} \\
% ~~~~~~BERT-large + NSPP    & 340M & 51.8 &  & n/a & n/a & n/a & 8.4  & 41.1 & 23.3 & 24.0 \\ 
%~~~~~~RoBERTa-large + NSPP & 355M & 67.3 &  & n/a & n/a & n/a & 33.5 & 64.6 & 46.3 & 50.2 \\

\cmidrule(lr){2-11}

~~~Our results \\
~~~~~~BERT-large (off-the-shelf) & 340M & 46.8          &               & 47.0 & 46.9 & 44.6 & 2.2 & 44.9 & 16.0 & 11.3 \\
~~~~~~~~~further pre-trained, NSPP with\\
~~~~~~~~~~~~affixal cues &      & \textbf{50.1} &\scriptsize($+$7.1)$^{*}$ & \textbf{49.3} & \textbf{51.3} & 47.8 & \textbf{6.7} & 46.0 & \textbf{20.1} & \textbf{19.8} \\
~~~~~~~~~~~~single-word cues &   & 48.3         &\scriptsize($+$3.3)$^{*}$  & 47.3 & 49.7          & 46.4          & 3.7 & \textbf{48.3} & 17.9 & 13.1 \\
~~~~~~~~~~~~multi-word cues &   & 48.9          &\scriptsize($+$4.4)$^\ast$  & 48.7 & 49.7          & 43.9          & 4.6 & 47.5 & 18.6 & 15.3 \\
~~~~~~~~~~~~All cues    &   & 48.6              &\scriptsize($+$3.8)$^{*}$  & 47.9 & 49.0 & \textbf{51.5} & 4.5 & 45.9 & 17.8 & 15.9 \\ \addlinespace

~~~~~~RoBERTa-large (off-the-shelf) & 355M & 54.1          &               & 51.3 & 57.2 & 52.9 &12.7 & 51.4 & 26.8 & 24.9 \\
~~~~~~~~~further pre-trained, NSPP with: \\
~~~~~~~~~~~~affixal cues &      & \textbf{68.0} &\scriptsize($+$25.8)$^{*}$ & \textbf{65.5} & \textbf{70.7} &\textbf{ 67.7} &\textbf{33.4} & \textbf{66.0} & \textbf{46.0} & \textbf{50.5} \\
~~~~~~~~~~~~single-word cues &   & 65.2          &\scriptsize($+$20.5)$^{*}$ & 63.1 & 67.5 & 64.9          & 30.7 & 63.6 & 43.5 & 47.4 \\
~~~~~~~~~~~~multi-word cues &   & 65.4          &\scriptsize($+$21.0)$^{*}$ & 63.1 & 68.2 & 63.5          & 29.1 & 60.6 & 41.6 & 46.6 \\
~~~~~~~~~~~~All cues    &   & 65.3          &\scriptsize($+$20.8)$^{*}$ & 63.9 & 66.8 & 65.6 & 29.1 & 62.3 & 41.9 & 46.4 \\

\midrule

LLMs, zero-shot \\
%~~~From~\citet{ravichander-etal-2022-condaqa}  \\
~~~Previous Work  \\
% ~~~~~~UnifiedQA-v2-11B & 11B & 73.1 & & n/a & n/a & n/a & 40.0 & 75.5 & 53.7 & 54.1 \\ % condaqa best results, it's an encoder-decoder Transformer not decoder-only.
% ~~~~~~GPT-3 & 175B & 43.7 & & n/a & n/a & n/a & 1.3 & 41.33 & 10.7 & 10.9 \\
~~~~~~text-davinci-002 [1] &  & 54.0 & & n/a & n/a & n/a & 16.3 & 55.5 & 29.9 & 27.8 \\

\cmidrule(lr){2-11}

~~~Our results  \\
~~~~~~Llama3.2-1B (off-the-shelf)           & 1B & 48.6 & & 48.0 & 48.9 & 50.7 & 2.1 & 52.3 & 17.7 & 10.2 \\
~~~~~~~~~further pre-trained, NSPP with: \\
~~~~~~~~~~~~affixal cues & & \textbf{50.1} &\scriptsize($+$3.0)$^{*}$ & \textbf{49.3} & \textbf{50.7} & \textbf{51.6} & \textbf{3.3} & \textbf{53.3} & \textbf{19.5} & \textbf{12.0} \\
~~~~~~~~~~~~single-word cues & & 48.8 &\scriptsize($+$0.3) & 48.3 & 49.3 & 48.4 & 1.8 & 52.4 & 17.7 & 10.3 \\
~~~~~~~~~~~~multi-word cues & & 49.3 &\scriptsize($+$1.4)$^{*}$ & 48.5 & 50.0 & 50.7 & 2.8 & 52.6 & 18.7 & 10.8 \\
~~~~~~~~~~~~All cues & & 48.8 &\scriptsize($+$0.3) & 48.4 & 48.9 & 50.2 & 2.3 & 52.4 & 17.7 & 10.3 \\

\addlinespace
~~~~~~Llama3.2-3B (off-the-shelf) & 3B & 58.9 & & 56.4 & 60.5 & 67.5 & 14.8 & 62.5 & 30.1 & 28.2 \\
~~~~~~~~~further pre-trained, NSPP with: \\
~~~~~~~~~~~~affixal cues    &&\textbf{59.6} &\scriptsize($+$1.1) & \textbf{56.6} & \textbf{61.9} & 66.9 & \textbf{15.3} & \textbf{65.1} & \textbf{30.6} & \textbf{28.7} \\
~~~~~~~~~~~~single-word cues && 59.1 &\scriptsize($+$0.3) & 55.9 & 61.6 & 66.8 & \textbf{15.3} & 64.6 & 30.2 & 28.4 \\
~~~~~~~~~~~~multi-word cues  && 59.3 &\scriptsize($+$0.7) & 56.3 & 61.6 & \textbf{69.3} & 15.2 & 64.9 & \textbf{30.6} & 28.0 \\
~~~~~~~~~~~~All cues    && 59.3 &\scriptsize($+$0.7) & 56.4 & 61.5 & 66.8 & 14.8 & 64.7 & 30.3 & 28.3 \\

% \addlinespace
% ~~~~~~Qwen2-0.5B (off-the-shelf) & 0.5B & 49.6 & & 48.4 & 50.9 & 49.6 & 3.8 & 47.0 & 18.4 & 15.2 \\
% ~~~~~~~~~further pre-trained, NSPP with: \\
% ~~~~~~~~~~~~affixal cues    && 49.9 &\scriptsize($+$0.7) & 49.6 & \textbf{50.7} & 46.2 & \textbf{5.1} & 45.1 & 17.4 & \textbf{17.4} \\
% ~~~~~~~~~~~~single-word cues && 48.4 &\scriptsize($-$2.5) & 48.1 & 49.0 & 44.2 & 4.1 & 44.9 & 16.1 & 15.6 \\
% ~~~~~~~~~~~~multi-word cues  && \textbf{50.0} &\scriptsize($+$0.8) & \textbf{49.8} & 50.2 & \textbf{48.9} & 4.7 & \textbf{46.1} & \textbf{17.6} & \textbf{17.4} \\
% ~~~~~~~~~~~~All cues    && 48.9 &\scriptsize($-$1.4) & 48.8 & 49.3 & 45.4 & 4.3 & 45.2 & 17.0 & 16.6 \\

\bottomrule
\end{tabular}

  \caption{
    Results on CondaQA with encoder-only LMs and decoder-only LLMs and zero-shot prompts.
    [1] refers to \citet{ravichander-etal-2022-condaqa},
    [2] to \citet{rezaei-blanco-2024-paraphrasing},
    and
    [3] to \citet{rezaei-blanco-2025-making}.
    Pars. denotes parameters; Par., Sco., and Aff. denote paraphrase, scope edit, and affirmative edit, respectively.
    $\Delta$ indicates the relative performance with respect to the off-the-shelf models,
    and 
    an asterisk ($^\ast$) indicates statistical significance for Accuracy (All) only (McNemar's test~\cite{mcnemar1947note}, $p<0.05$).
    % an asterisk ($^\ast$) indicates a statistically significant difference (McNemar's test~\cite{mcnemar1947note}, $p<0.05$).
    Both LMs and LLMs benefit from further NSPP pre-training over their off-the-shelf counterparts, especially with affixal negation cues.
    }
  \label{t:condaqa_main_results}
\end{table*}

\begin{table*}
  \centering
  \small

\begin{tabular}{l r@{ }l r@{ }l r@{ }l r@{ }l rrrr}
\toprule
& \multicolumn{8}{c}{\textbf{Accuracy}} & \multicolumn{4}{c}{\textbf{Group Consistency}} \\
\cmidrule(lr){2-9} \cmidrule(lr){10-13}
& $Q_1$ &($\Delta$\%) & $Q_2$ &($\Delta$\%) & $Q_3$ &($\Delta$\%) & $Q_4$ &($\Delta$\%) & $Q_1$ & $Q_2$ & $Q_3$ & $Q_4$  \\
\midrule

BERT-large
& 46.7 & & 47.1 & & 45.9 & & 48.8 & & 2.5  & 1.7 & 1.8  & 0.0 \\
~~~+ NSPP, affixal
& 50.0 &\scriptsize$(+7.1)$ & 50.3 &\scriptsize$(+6.8)$ & 49.5 &\scriptsize$(+7.8)$ & 52.2 &\scriptsize$(+7.0)$ & 7.4  & 4.6 & 7.1  & 2.0 \\

RoBERTa-large
& 55.3 & & 52.0 && 48.6 && 51.4 && 14.3 & 8.0  & 8.8  & 12.0 \\
~~~+ NSPP, affixal
& 68.7 &\scriptsize$(+24.2)$ & 67.4 &\scriptsize$(+29.6)$ & 66.6 &\scriptsize$(+37.0)$ & 62.1 &\scriptsize$(+20.8)$ & 34.5 & 32.1 & 31.0 & 22.0 \\

\addlinespace
Llama3.2-1B 
& 48.9 && 48.3 && 47.4 && 48.0 && 2.3  & 2.6  & 0.9  & 0.0 \\
~~~+ NSPP, affixal
& 50.3 &\scriptsize$(+2.9)$ & 49.5 &\scriptsize$(+2.5)$ & 49.3 &\scriptsize$(+4.0)$ & 49.1 &\scriptsize$(+2.3)$ & 3.4  & 3.4  & 1.7  & 2.0 \\

Llama3.2-3B 
& 59.5 && 58.3 && 58.2 && 54.6 && 16.1  & 12.8  & 10.4  & 8.0 \\
~~~+ NSPP, affixal
& 60.3 &\scriptsize$(+1.3)$ & 59.1 &\scriptsize$(+1.4)$ & 56.4 &\scriptsize$(-3.1)$ & 53.3 &\scriptsize$(-2.4)$ & 16.5  & 14.0  & 8.7  & 8.0 \\

% Qwen2-0.5B 
% & 50.2 && 50.0 && 46.1 && 45.7 && 4.9  & 2.1  & 2.6  & 0.0 \\
% ~~~+ NSPP, multi
% & 50.5 &\scriptsize$(+0.6)$ & 48.5 &\scriptsize$(-3.0)$ & 48.9 &\scriptsize$(+6.1)$ & 47.5 &\scriptsize$(+3.9)$ & 4.6  & 3.4  & 6.1  & 2.0 \\

\bottomrule
\end{tabular}

  \caption{
      % Results with the best performing NSPP pre-training per model~(Table \ref{t:condaqa_main_results}) on four subsets of CondaQA
      % containing cues based on frequency
      % (four quartiles; $Q_1$ contains the top-25\% most frequent cues).
      % $\Delta$ indicates the relative change with respect to the off-the-shelf model.
      % Pre-training with NSPP is beneficial regardless of cue frequency with both LMs and LLMs.
      % In particular, we observe substantial gains in consistency.
      % Bold values indicate the best result for each model.
      Results with the best NSPP pre-training per model (Table~\ref{t:condaqa_main_results}) on four CondaQA subsets grouped by cue frequency (quartiles; $Q_1$ contains the top-25\% most frequent cues). 
      $\Delta$ indicates relative change with respect to the off-the-shelf model. 
      % NSPP pre-training improves performance across all frequency groups for both LMs and LLMs.
      NSPP pre-training yields stronger performance gains on moderately frequent cues ($Q_2$, $Q_3$).
      %, with particularly large gains in consistency.
    }
  \label{t:condaqa_freq_results}
\end{table*}
We first discuss the results with the largest benchmark, CondaQA,
and then with the remaining benchmarks.
Pre-training both encoder-only LMs and smaller LLMs for NSPP with NegCue
consistently yields better results across all benchmarks.

\subsection{Results on Large-Scale CondaQA} 
% We analyze results on CondaQA .
% We first analyze the results on CondaQA (Table~\ref{t:condaqa_main_results}).

% \noindent
\textbf{Pre-training with NegCue benefits both LMs and LLMs.}
As shown in Table~\ref{t:condaqa_main_results} (Accuracy, All),
both LMs and LLMs further pre-trained with NSPP outperform their off-the-shelf counterparts 
(BERT: +3.3--7.1\%;
RoBERTa: +20.5--25.8\%;
Llama3.2-1B/3B: +0.3--3.0\% / +0.7--1.1\%, depending on the specific NegCue subset).
% Except for Qwen
% (+0.7--0.8\%), the differences with respect to the off-the-shelf models are statistically significant,
% with affixal pre-training yielding the strongest performance.
Our results with RoBERTa and NSPP pre-training with affixal negation (Acc.: 68\%)
outperform previous work \cite{rezaei-blanco-2025-making} targeting only three frequent negation cues (\emph{not}, \emph{n't}, and \emph{never}). 
% Further, we surpass 67.1\% achieved with affirmative interpretations without additional computation during inference.
Further, we surpass the 67.1\% accuracy achieved using affirmative interpretations, without additional computation during inference.
%while requiring neither additional computation nor doubled input size during inference.
% Lastly, we also outperform UnifiedQA-v2-large (66.7\%) despite having twice the parameters and being fine-tuned with 1M question-answer pairs.
Lastly, we also outperform UnifiedQA-v2-large (66.7\%), which has twice as many parameters and is fine-tuned on 1M question-answer pairs.
For decoder-only LLMs, both Llama model sizes consistently improve after further pre-training.
Despite its much smaller scale, the 1B model narrows the gap to the closed-source text-davinci-002 model, highlighting the value of further pre-training for small LLMs.

% \noindent
\textbf{Affixal negation yields the largest gains, though all types remain beneficial.}
% We further analyze performance by negation cue type.
Across LMs and LLMs,
pre-training on affixal and multi-word cues yields larger improvements over off-the-shelf models
while gains from single-word pre-training are consistently smaller.
This pattern suggests that models may already handle single-word negation reasonably well, whereas affixal and multi-word negation remain more challenging and therefore benefit more from targeted pre-training. 
This interpretation is supported by the cue-type analysis on CondaQA (Table~\ref{t:condaqa_main_results}, Columns~4--6). 
Accuracy is generally highest on the single-word subset across models, indicating stronger baseline performance. 
We believe this is one reason why affixal negation yields the strongest generalization across negation cue types, 
as it is less common and less studied than single-word negation.
In addition, affixal negation contains a more balanced and diverse set of unique cues (Figure~\ref{fig.nspp_cue_donut}, Table~\ref{t:nspp_corpora}).
By contrast, the single-word and multi-word categories are heavily dominated by a few high-frequency cues.
This imbalance also carries over to the \textit{All Cues} subset, which may explain why training on it does not produce the largest gains despite covering all types.

% \noindent
\textbf{Moderately frequent negation cues benefit more from pre-training.} 
We further analyze results on CondaQA by negation cue frequency (Table~\ref{t:condaqa_freq_results}).
Specifically, we create four partitions of the test split depending on cue frequency
(four quartiles, $Q_1$–$Q_4$, where $Q_1$ contains the top-25\% most frequent cues).
Across models, the largest improvements ($\Delta$\%)  generally occur in $Q_2$ and $Q_3$, particularly $Q_3$, while performance gradually decreases from $Q_1$ to $Q_4$. 
This is consistent with our earlier findings on negation types: extremely common cues in $Q_1$ exhibit strong baseline performance, 
whereas extremely rare cues in $Q_4$ receive relatively little exposure during pre-training due to their low frequency. % in NegCue.
%both the evaluation benchmarks and NegCue. 
As a result, the largest gains emerge in the middle-frequency groups, where cues remain sufficiently challenging while still appearing often enough to benefit from additional training. 
% Nevertheless, NSPP pre-training generally yields improvements across all cue-frequency groups.
Nevertheless, NSPP pre-training generally yields positive gains across cue-frequency groups.

% Across off-the-shelf models,
% accuracy and group consistency tend to be higher on subsets containing more frequent cues,
% indicating that commonly occurring cues are easier for off-the-shelf models.
% Pre-training with NSPP results in consistent improvements regardless of cue frequency.
% We note that the relative improvements of pre-training ($\Delta$\%) are often larger for
% the 50\% least frequent negation cues~($Q_3$ and $Q_4$).
% In other words, NSPP pre-training benefits all negation cues at test time,
% but especially the infrequent ones.

\begin{table*}[t!]
  \centering
  \small
  %\resizebox{\linewidth}{!}{
  \begin{tabular}{l r r@{ }l r@{ }l r@{ }l r@{ }l}
\toprule
           && \multicolumn{2}{l}{\textbf{NeQA}} &
              \multicolumn{2}{l}{\textbf{NevIR}} &
              \multicolumn{2}{l}{\textbf{NMoNLI}} &
              \multicolumn{2}{l}{\textbf{ScoNe-NLI}} \\ 
              % \cmidrule(lr){3-10}
              \midrule
    
& \# Pars. & Acc. &($\Delta$\%) 
             & Acc. &($\Delta$\%) 
             & Acc. &($\Delta$\%) 
             & Acc. &($\Delta$\%) \\ 
             \midrule

Fine-tuned encoder-only LMs  \\
~~~Previous Work \\
~~~~~~BERT-SNLI [1] 
&       & n/a & & n/a & &2.2   & & n/a & \\ 
% ~~~From~\citet{she-etal-2023-scone}  \\ % ScoNe paper
~~~~~~RoBERTa-large-MAF-NLI [2] 
& 355M  & n/a & & n/a & &n/a   & & 65.8 & \\ 
% fintuned on MAF-NLI, stands for on MNLI, ANLI, and Fever-NLI.
%~~~~~~stsb-roberta-large \cite{weller-etal-2024-nevir} 
% & 355M & n/a &  & 24.9 & & n/a & & n/a &  \\
% ~~~~~~MonoT5 3B \cite{nogueira-etal-2020-document}     
% & 3B   & n/a &  & 50.6 & & n/a & & n/a &  \\ % best result in NevIR paper

\cmidrule(lr){2-10}

~~~Our results \\
% Our LMs results
~~~~~~BERT-large (off-the-shelf) & 340M 
                     & 70.0 &               
                     & 69.7 & 
                     & 6.0  & 
                     & 62.8 & \\
~~~~~~~~~further pre-trained, NSPP with \\
~~~~~~~~~~~~affixal cues &      
                     & \textbf{83.0} &\scriptsize$(+18.6)$$^\ast$  
                     & \textbf{72.0} &\scriptsize$(+3.3)$$^\ast$ 
                     & 8.0  &\scriptsize$(+33.3)$ 
                     & 63.3 &\scriptsize$(+0.8)$  \\
~~~~~~~~~~~~single-word cues &   
                     & 80.0 &\scriptsize$(+14.3)$  
                     & 71.1 &\scriptsize$(+2.0)$          
                     & 6.0 &\scriptsize$(+0.0)$ 
                     & \textbf{64.6} &\scriptsize$(+2.9)$$^\ast$ \\
~~~~~~~~~~~~multi-word cues &   
                     & 81.0 &\scriptsize$(+15.7)$  
                     & 71.8 &\scriptsize$(+3.0)$$^\ast$          
                     & 7.5  &\scriptsize$(+25.0)$
                     & \textbf{64.6} &\scriptsize$(+2.9)$$^\ast$ \\
~~~~~~~~~~~~All cues    &   
                     & 78.0 &\scriptsize$(+11.4)$  
                     & 71.5 &\scriptsize$(+2.6)$$^\ast$ 
                     & \textbf{19.5} &\scriptsize$(+225.0)$$^\ast$ 
                     & 64.5 &\scriptsize$(+2.7)$ \\
\addlinespace

~~~~~~RoBERTa-large (off-the-shelf)    & 355M 
                     & 42.0 & 
                     & 73.8 & 
                     & 6.0 & 
                     & 64.4 & \\
~~~~~~~~~further pre-trained, NSPP with \\
~~~~~~~~~~~~affixal cues    &
                     & 44.0 &\scriptsize$(+4.8)$ 
                     & \textbf{77.4} &\scriptsize$(+5.0)$$^\ast$ 
                     & 14.0 &\scriptsize$(+133.3)$$^\ast$ 
                     & 61.7 &\scriptsize$(-4.2)$ \\
% ~~~~~~~~~old Single-word &
%                      & 48.0 &$(+14.3)$ 
%                      & 61.5 &$(+1.3)$  
%                      & 6.5  &$(+8.3)$
%                      & \textbf{65.2} &$(+1.2)$ \\
~~~~~~~~~~~~single-word cues &
                     & 45.0 &\scriptsize$(+7.1)$ 
                     & 74.7 &\scriptsize$(+1.3)$  
                     & 7.0  &\scriptsize$(+16.7)$
                     & 64.3 &\scriptsize$(-0.2)$ \\
~~~~~~~~~~~~multi-word  cues&  
                     & \textbf{51.0} &\scriptsize$(+21.4)$ 
                     & 76.7 &\scriptsize$(+4.0)$$^\ast$ 
                     & \textbf{21.5} &\scriptsize$(+258.3)$$^\ast$ 
                     & \textbf{64.6} &\scriptsize$(+0.3)$  \\
~~~~~~~~~~~~All cues    &   
                     & \textbf{51.0} &\scriptsize$(+21.4)$ 
                     & 76.6 &\scriptsize$(+3.9)$$^\ast$ 
                     & 4.5  &\scriptsize$(-25.0)$ 
                     & 64.3 &\scriptsize$(-0.2)$ \\

\midrule

LLMs, zero-shot \\
~~~Previous Work \\
~~~~~~Cohere-medium [3]  & 6.1B & 44.0 & & n/a & &n/a   & & n/a & \\
%~~~~~~From~\citet{zhang-etal-2023-beyond}  \\ % NeQA paper
~~~~~~text-davinci-003 [2][3] &  & 71.0 & & n/a & &n/a   & & 64.0 & \\ % neqa best result 

% ~~~Previous Work \\ % version 1:
% %~~~~~~From~\citet{zhang-etal-2023-beyond}  \\ % NeQA paper
% ~~~~~~text-davinci-003 \cite{zhang-etal-2023-beyond} & n/a  & 71.0 & & n/a & &n/a   & & 64.0 & \\ % neqa best result 
% ~~~~~~Cohere-medium  \cite{zhang-etal-2023-beyond}  & 6.1B & 44.0 & & n/a & &n/a   & & n/a & \\
% %~~~From~\citet{she-etal-2023-scone}  \\ % ScoNe paper
% ~~~~~~text-davinci-002 \cite{she-etal-2023-scone} & n/a  & n/a  & & n/a & &n/a   & & 66.0 & \\ 

\cmidrule(lr){2-10}

~~~Our results \\
~~~~~~Llama3.2-1B  (off-the-shelf)     & 1B 
                     & 38.0 & 
                     & 43.1  & 
                     & 43.5 & 
                     & 49.2 & \\
~~~~~~~~~further pre-trained, NSPP with \\
~~~~~~~~~~~~affixal cues    & 
                     & 40.0 &\scriptsize$(+5.3)$ 
                     & 50.6 &\scriptsize$(+17.5)$$^\ast$
                     & \textbf{51.0} &\scriptsize$(+17.2)$ 
                     & \textbf{51.2} &\scriptsize$(+4.1)$$^\ast$ \\
~~~~~~~~~~~~single-word cues & 
                     & 39.0 &\scriptsize$(+2.6)$ 
                     & 50.8  &\scriptsize$(+17.9)$$^\ast$ 
                     & 50.0 &\scriptsize$(+14.9)$ 
                     & 49.8 &\scriptsize$(+1.2)$ \\
~~~~~~~~~~~~multi-word cues & 
                     & \textbf{45.0} &\scriptsize$(+18.4)$ 
                     & 49.4 &\scriptsize$(+14.6)$$^\ast$ 
                     & \textbf{51.0} &\scriptsize$(+17.2)$
                     & \textbf{51.2} &\scriptsize$(+4.1)$$^\ast$ \\
~~~~~~~~~~~~All cues    & 
                     & 38.0 &\scriptsize$(+0.0)$ 
                     & \textbf{51.3} &\scriptsize$(+19.1)$$^\ast$
                     & 50.0 &\scriptsize$(+14.9)$
                     & 49.7 &\scriptsize$(+1.0)$\\

\addlinespace
~~~~~~Llama3.2-3B   (off-the-shelf)     & 3B 
                     & 63.0 & 
                     & 47.8  & 
                     & 48.5 & 
                     & 53.4 & \\
~~~~~~~~further pre-trained, NSPP with \\
~~~~~~~~~~~~affixal cues    &
                     & 70.0 &\scriptsize$(+11.1)$$^\ast$ 
                     & 51.3 &\scriptsize$(+7.3)$$^\ast$
                     & \textbf{50.0} &\scriptsize$(+3.1)$ 
                     & \textbf{64.4} &\scriptsize$(+20.6)$$^\ast$ \\
~~~~~~~~~~~~single-word cues&
                     & 69.0 &\scriptsize$(+9.5)$$^\ast$ 
                     & 51.3  &\scriptsize$(+7.3)$$^\ast$ 
                     & 49.5 &\scriptsize$(+2.1)$ 
                     & 61.4 &\scriptsize$(+15.0)$$^\ast$ \\
~~~~~~~~~~~~multi-word cues &
                     & \textbf{73.0} &\scriptsize$(+15.9)$$^\ast$ 
                     & 51.5 &\scriptsize$(+7.7)$$^\ast$  
                     & 49.0 &\scriptsize$(+1.0)$ 
                     & 63.8 &\scriptsize$(+19.5)$$^\ast$ \\
~~~~~~~~~~~~All cues    &
                     & 68.0 &\scriptsize$(+7.9)$ 
                     & \textbf{51.7}  &\scriptsize$(+8.3)$$^\ast$  
                     & 49.0 &\scriptsize$(+1.0)$ 
                     & 63.1 &\scriptsize $(+18.2)$$^\ast$ \\
\bottomrule
\end{tabular}

  %}
  \caption{
      % Results on four benchmarks involving negation cues under two evaluation settings: fine-tuned LMs (top) and zero-shot LLMs (bottom).
      % The first block in each section shows results from prior work. 
      % The best result for each model is in bold and Delta ($\Delta$) indicates the percentage change relative to the off-the-shelf model.
      Results on four NLU benchmarks. 
      Same notation as Table~\ref{t:condaqa_main_results}.
      [1]:~\citet{geiger-etal-2020-neural} (BERT fine-tuned on SNLI); [2]:~\citet{she-etal-2023-scone} (fine-tuned on the MAF-NLI datasets); 
      [3]:~\citet{zhang-etal-2023-beyond}. 
      ``n/a'' indicates that comparable baselines were not reported.
      % $\Delta$ indicates relative improvements with respect to the off-the-shelf model.
      % Both LMs and LLMs further pre-trained with NegCue, yield better results than their off-the-shelf counterparts across the four corpora.
      Further pre-training with NegCue yields better results across the four corpora.
      We also outperform results by previous work (when available) by substantial margins despite experimenting with much smaller models (e.g., on NeQA, off-the-shelf Llama3.2-1B achieves 38\% accuracy, substantially below Cohere-medium (44\%), 
      while our further pre-trained version reaches 45\% despite being  six times smaller.).
    }
  \label{t:other4benchmarks_results}
\end{table*}

\subsection{Results on other NLU Benchmarks}
% We next discuss results on the remaining four NLU benchmarks (Table~\ref{t:other4benchmarks_results}).
We next discuss results on the remaining four NLU benchmarks, reported in Table~\ref{t:other4benchmarks_results}.
% Both LMs and LLMs pre-trained on affixal or multi-word cues generally achieve the best performance.
% This trend aligns with our findings on CondaQA, indicating that targeted exposure to these cue types is beneficial. % for negation understanding.

% \noindent
\textbf{NeQA: further pre-training surpasses previously reported results.} 
Interestingly, BERT-large outperforms RoBERTa-large on NeQA (83\% vs. 51\%),
despite the latter being a stronger model in many settings, consistent with the benchmark's reported inverse scaling behavior. 
For LLMs, the further pre-trained 1B model outperforms the much larger Cohere-medium (45\% vs.\ 44\%), while the best 3B model achieves 73\%, surpassing the previous best result of 71\% on this benchmark. 
This suggests that negation-focused pre-training can be more effective than model scaling alone. 
% Overall, both LMs and LLMs benefit from further pre-training on our corpora.

% \noindent
% \textbf{NevIR: both LMs and LLMs achieve statistically significant improvements.} 
\textbf{NevIR: Pretraining with NegCue achieves statistically significant improvements with both LMs and LLMs.}
Although no directly comparable baselines have been reported, our results show that pre-training on our corpora leads to clear gains on this negation-sensitive task. 
Compared to their off-the-shelf counterparts, encoder-only LMs achieve improvements of 1.3--4.9\%, while LLMs improve by 7.3--19.1\%. 
% Notably, encoder-only LMs fine-tuned on the task-specific training set outperform smaller LLMs evaluated in a zero-shot setting.
Notably, encoder-only LMs fine-tuned on the task-specific training set outperform smaller LLMs evaluated zero-shot.

% The results show that further pre-training on our corpora leads to clear gains on negation-sensitive information retrieval tasks.
% Across encoder-only LMs, all further pre-trained variants outperform their off-the-shelf counterparts
% (BERT: +2.3--3.3\%,
% RoBERTa: +1.3--3.5\%).
% For LLMs, performance varies across models.
% Qwen2-0.5B shows limited benefits from further pre-training under some cue settings (-10.5--+11.6\%),
% whereas Llama3.2-1B benefits substantially, achieving improvements over 200\%.

% \noindent
\textbf{NLI: pre-training with NegCue benefits both LMs and LLMs.} 
NMoNLI and ScoNe-NLI require deeper negation understanding, including negation scope reasoning. 
Nearly all models improve after NSPP pre-training on our corpora, with only a few RoBERTa-large variants showing no gains. 
Pre-trained LLMs improve on both benchmarks, especially on ScoNe-NLI, while LMs show larger gains on NMoNLI. 
Notably, RoBERTa-large pre-trained on multi-word cues improves by more than 200\%. 
Although performance on NMoNLI remains relatively low, the large relative gains highlight the effectiveness of our pre-training approach. % on this challenging task.

% Across both benchmarks, further pre-training consistently benefits both LMs and LLMs:
% the best-performing variant outperforms the off-the-shelf counterparts.
% Improvements are substantially larger on NMoNLI than on ScoNe-NLI, with both LMs achieving gains exceeding 200\% on NMoNLI.
% In contrast, Qwen2-0.5B shows limited benefits from further pre-training on both NLI benchmarks.
% We observe that this model often predicts a single label across many instances, a behavior that is already present in its off-the-shelf version, suggesting that these tasks are particularly challenging for this model.

%%%%%%%%%%%%%%%%%%%%%%%%%%%%%%%%%%%%%%%%%%%%%%%%%%%%%%%%%%%%%%%%%%%%%%%%%%%%%%%%
%%%%%%%%%%%%%%%%%%%%%%%%%%%%%%%%%%%%%%%%%%%%%%%%%%%%%%%%%%%%%%%%%%%%%%%%%%%%%%%%
% added after rebattul
\section{Ablation Studies}
\label{s:ablation}
%%%%%%%%%%%%%%%%%%%%%%%%%%%%%%%%%%%%%%%%%%%%%%%%%%%%%%%%%%%%%%%%%%%%%%%%%%%%%%%%
%%%%%%%%%%%%%%%%%%%%%%%%%%%%%%%%%%%%%%%%%%%%%%%%%%%%%%%%%%%%%%%%%%%%%%%%%%%%%%%%
To further validate our findings, we conduct two ablation studies on CondaQA and NeQA. 
First, we test whether the observed gains arise from negation-specific learning or simply from generic continued pre-training effects. 
Second, to better isolate the effect of affixal negation, we construct balanced corpora with an equal number of unique negation cues across cue types and the same number of samples for each cue.
% Additional details are provided in the Appendix.

\begin{table*}[t]
  \centering
  \small
  \begin{tabular}{llcccc}
\toprule
\textbf{Benchmark} & \textbf{Setting} & \textbf{BERT-Large} & \textbf{RoBERTa-Large} & \textbf{Llama3.2-1B} & \textbf{Llama3.2-3B} \\
\midrule

\multirow{5}{*}{CondaQA}
& Off-the-shelf & 46.8 & 54.1 & 48.6 & 58.9 \\
& NSP & 46.0 & 63.9 & 32.3 & 51.8 \\
& NSPP, affixal & \textbf{50.1} & \textbf{68.0} & \textbf{50.1} & \textbf{59.6} \\
& NSPP, single & 48.3 & 65.2 & 48.8 & 59.1 \\
& NSPP, multi & 48.9 & 65.2 & 49.3 & 59.3 \\
\midrule

\multirow{5}{*}{NeQA}
& Off-the-shelf & 70.0 & 42.0 & 38.0 & 63.0 \\
& NSP & 23.0 & 37.0 & 31.0 & 67.0 \\
& NSPP, affixal & \textbf{83.0} & 44.0 & 40.0 & 70.0 \\
& NSPP, single & 80.0 & 45.0 & 39.0 & 69.0 \\
& NSPP, multi & 81.0 & \textbf{51.0} & \textbf{45.0} & \textbf{73.0} \\
\bottomrule
\end{tabular}
  \caption{
            Performance on CondaQA and NeQA after continued pre-training with NSP or NSPP using different negation cue types. 
            The reported metric is accuracy, and the best results are shown in bold.
            %Additional NSP continued pre-training generally underperforms the corresponding off-the-shelf models.
        }
  \label{t:nsp_ablation_results}
\end{table*}

\begin{table*}[t]
  \centering
  \small
  \begin{tabular}{llcccc}
\toprule
\textbf{Benchmark} & \textbf{Setting} & \textbf{BERT-Large} & \textbf{RoBERTa-Large} & \textbf{Llama3.2-1B} & \textbf{Llama3.2-3B} \\
\midrule

\multirow{3}{*}{CondaQA}
& NSPP, affixal & \textbf{47.9} & \textbf{66.4} & \textbf{52.3} & \textbf{59.6} \\
& NSPP, single  & 44.3 & 66.1 & 49.3 & 55.6 \\
& NSPP, multi   & 46.0 & 65.8 & 52.0 & 59.5 \\
\midrule

\multirow{3}{*}{NeQA}
& NSPP, affixal & \textbf{82.0} & \textbf{55.0} & \textbf{46.0} & \textbf{63.0} \\
& NSPP, single  & 79.0 & 54.0 & 38.0 & 62.0 \\
& NSPP, multi   & 74.0 & 52.0 & 45.0 & \textbf{63.0} \\
\bottomrule
\end{tabular}
  \caption{
            Performance on CondaQA and NeQA after NSPP continued pre-training on the balanced affixal, single-word, and multi-word corpora.
            We report accuracy, with the best results shown in bold.
  }
  \label{t:balanced_ablation_results}
\end{table*}

%%%%%%%%% NSP continued pre-training %%%%%%%%%%
\textbf{Generic continued pre-training does not improve performance and instead leads to substantial degradation.} % on non-negated data 
To verify that the observed gains are not simply due to additional pre-training, we construct a corpus from affirmative samples in NegCue, matching the size of each NegCue subset (affixal, single-word, and multi-word). 
We then perform continued pre-training with the standard next sentence prediction (NSP) objective~\cite{devlin-etal-2019-bert}, 
while keeping all other experimental settings identical to those used for NSPP pre-training. 
% As shown in Table~\ref{t:nsp_ablation_results}, non-negated continued pre-training substantially degrades performance, often falling below the corresponding off-the-shelf models. 
As shown in Table~\ref{t:nsp_ablation_results}, NSP-based continued pre-training substantially degrades performance, often falling below the corresponding off-the-shelf models. 
%It also consistently underperforms all NegCue-based variants across single-word, multi-word, and affixal negation. 
It also consistently underperforms all NegCue-based variants across the three negation types.
These results provide evidence that the gains in negation understanding arise from negation-specific learning with NegCue rather than generic continued pre-training effects.
% As shown in Table~\ref{t:nsp_ablation_results}, we make two main observations. 
% First, non-negated continued pre-training leads to substantial performance degradation, with the resulting models often performing worse than their off-the-shelf counterparts. 
% Second, models obtained through non-negated continued pre-training consistently underperform all NegCue-based continued pre-training variants, including those based on single-word, multi-word, and affixal negation. 

%%%%%%%%% isolate affixal %%%%%%%%%%
\textbf{Affixal negation remains the most effective negation type across all models.}
%even when the number of unique cues is controlled.}
As discussed in Section~\ref{s:datasets_framework}, the affixal subset of NegCue contains the largest number of unique cues, 
and the original corpus follows the natural cue distribution, resulting in varying numbers of samples across cues.
To further isolate the effect of negation type, we construct three fully balanced corpora.
Specifically, from each of the affixal, single-word, and multi-word NegCue subsets, 
we select the six most frequent unique cues and draw an equal number of samples for each cue (10,476).
This results in three corpora with identical sizes and the same number of samples per cue, each containing 62,856 training, 30,984 validation, and 30,988 test samples.
We then continue pre-training the models with NSPP on each of the three corpora under identical settings and evaluate them on CondaQA and NeQA. 
As shown in Table~\ref{t:balanced_ablation_results}, further pre-training on the affixal corpus consistently achieves the best performance under this controlled setting, 
outperforming both the single-word and multi-word corpora across all models. 
In contrast, single-word continued pre-training generally yields the lowest performance, consistent with our main findings.
%%%%%%%%%%%%%%%%%%%%%%%%%%%%%%%%%%%%%%%%%%%%%%%%%%%%%%%%%%%%%%%%%%%%%%%%%%%%%%%%
%%%%%%%%%%%%%%%%%%%%%%%%%%%%%%%%%%%%%%%%%%%%%%%%%%%%%%%%%%%%%%%%%%%%%%%%%%%%%%%%
\section{Conclusion}
\label{s:conclusions}
%%%%%%%%%%%%%%%%%%%%%%%%%%%%%%%%%%%%%%%%%%%%%%%%%%%%%%%%%%%%%%%%%%%%%%%%%%%%%%%%
%%%%%%%%%%%%%%%%%%%%%%%%%%%%%%%%%%%%%%%%%%%%%%%%%%%%%%%%%%%%%%%%%%%%%%%%%%%%%%%%

Negation remains a challenging problem for language models, while prior work often focuses on only a few highly frequent negation cues, such as \textit{not} and \textit{never}.
% We construct NegCue, a large-scale negation corpus covering over 200 negation cues, % beyond these commonly studied forms, 
% enabling the study of how different negation types affect negation understanding.
We construct NegCue, a large-scale negation corpus covering over 200 negation cues spanning three negation types (affixal, single-word, and multi-word), enabling the study of how different negation types affect negation understanding.
We further pre-train both encoder-only LMs and LLMs on Next Sentence Polarity Prediction, a negation-focused task. 
Experiments on five benchmarks show consistent improvements. 
Notably, affixal negation yields substantially larger gains, while the commonly studied single-word negation provides only modest improvements.
These findings suggest that current models exhibit uneven understanding across negation types, 
highlighting promising directions for improving negation understanding and broader language understanding.

\section*{Limitations}
This work focuses on diverse negation types and cues beyond the small set of commonly studied forms.
Our analysis, however, is limited to English corpora, and extending it to other languages with different negation systems remains future work. 
While the negation cues within each negation type in NegCue follow their natural corpus distributions, some of our findings suggest that such imbalance may limit the effectiveness of pre-training. 
In particular, naturally distributed corpora may become dominated by a few high-frequency cues while providing limited exposure to extremely rare cues. 
Further exploring alternative balancing strategies may therefore help better understand the role of cue distributions in negation-focused pre-training. 
In addition, our findings suggest that different negation cues contribute unequally to negation understanding, raising an interesting direction for identifying which cues are most beneficial for improving language models. 
Finally, while a simple zero-shot prompting strategy already yields promising results, it would be worthwhile to investigate few-shot and alternative prompting strategies.

\bibliography{custom}

@inproceedings{truong-etal-2024-revisiting,
    title = "Revisiting subword tokenization: A case study on affixal negation in large language models",
    author = "Truong, Thinh Hung  and
      Otmakhova, Yulia  and
      Verspoor, Karin  and
      Cohn, Trevor  and
      Baldwin, Timothy",
    editor = "Duh, Kevin  and
      Gomez, Helena  and
      Bethard, Steven",
    booktitle = "Proceedings of the 2024 Conference of the North American Chapter of the Association for Computational Linguistics: Human Language Technologies (Volume 1: Long Papers)",
    month = jun,
    year = "2024",
    address = "Mexico City, Mexico",
    publisher = "Association for Computational Linguistics",
    url = "https://aclanthology.org/2024.naacl-long.284/",
    doi = "10.18653/v1/2024.naacl-long.284",
    pages = "5082--5095"
}

@inproceedings{ye-etal-2023-assessing,
    title = "Assessing Step-by-Step Reasoning against Lexical Negation: A Case Study on Syllogism",
    author = "Ye, Mengyu  and
      Kuribayashi, Tatsuki  and
      Suzuki, Jun  and
      Kobayashi, Goro  and
      Funayama, Hiroaki",
    editor = "Bouamor, Houda  and
      Pino, Juan  and
      Bali, Kalika",
    booktitle = "Proceedings of the 2023 Conference on Empirical Methods in Natural Language Processing",
    month = dec,
    year = "2023",
    address = "Singapore",
    publisher = "Association for Computational Linguistics",
    url = "https://aclanthology.org/2023.emnlp-main.912/",
    doi = "10.18653/v1/2023.emnlp-main.912",
    pages = "14753--14773"
}

@inproceedings{truong-etal-2022-another,
    title = "Not another Negation Benchmark: The {N}a{N}-{NLI} Test Suite for Sub-clausal Negation",
    author = "Truong, Hung Thinh  and
      Otmakhova, Yulia  and
      Baldwin, Timothy  and
      Cohn, Trevor  and
      Lau, Jey Han  and
      Verspoor, Karin",
    editor = "He, Yulan  and
      Ji, Heng  and
      Li, Sujian  and
      Liu, Yang  and
      Chang, Chua-Hui",
    booktitle = "Proceedings of the 2nd Conference of the Asia-Pacific Chapter of the Association for Computational Linguistics and the 12th International Joint Conference on Natural Language Processing (Volume 1: Long Papers)",
    month = nov,
    year = "2022",
    address = "Online only",
    publisher = "Association for Computational Linguistics",
    url = "https://aclanthology.org/2022.aacl-main.65/",
    doi = "10.18653/v1/2022.aacl-main.65",
    pages = "883--894"
}

@inproceedings{williams-etal-2018-broad,
    title = "A Broad-Coverage Challenge Corpus for Sentence Understanding through Inference",
    author = "Williams, Adina  and
      Nangia, Nikita  and
      Bowman, Samuel R.",
    editor = "Walker, Marilyn  and
      Ji, Heng  and
      Stent, Amanda",
    booktitle = "Proceedings of the 2018 Conference of the North {A}merican Chapter of the Association for Computational Linguistics: Human Language Technologies, Volume 1 (Long Papers)",
    month = jun,
    year = "2018",
    address = "New Orleans, Louisiana",
    publisher = "Association for Computational Linguistics",
    url = "https://aclanthology.org/N18-1101/",
    doi = "10.18653/v1/N18-1101",
    pages = "1112--1122"
}

@inproceedings{truong-etal-2022-improving,
    title = "Improving negation detection with negation-focused pre-training",
    author = "Truong, Hung Thinh  and
      Baldwin, Timothy  and
      Cohn, Trevor  and
      Verspoor, Karin",
    editor = "Carpuat, Marine  and
      de Marneffe, Marie-Catherine  and
      Meza Ruiz, Ivan Vladimir",
    booktitle = "Proceedings of the 2022 Conference of the North American Chapter of the Association for Computational Linguistics: Human Language Technologies",
    month = jul,
    year = "2022",
    address = "Seattle, United States",
    publisher = "Association for Computational Linguistics",
    url = "https://aclanthology.org/2022.naacl-main.309/",
    doi = "10.18653/v1/2022.naacl-main.309",
    pages = "4188--4193"
}

@inproceedings{varshney2025investigating,
  title={Investigating and addressing hallucinations of llms in tasks involving negation},
  author={Varshney, Neeraj and Raj, Satyam and Mishra, Venkatesh and Chatterjee, Agneet and Saeidi, Amir and Sarkar, Ritika and Baral, Chitta},
  booktitle={Proceedings of the 5th workshop on trustworthy NLP (TrustNLP 2025)},
  pages={580--598},
  year={2025}
}

@inproceedings{kim2025semantic,
  title={Semantic Inversion, Identical Replies: Revisiting Negation Blindness in Large Language Models},
  author={Kim, Jinsung and Koo, Seonmin and Lim, Heui-Seok},
  booktitle={Proceedings of the 2025 Conference on Empirical Methods in Natural Language Processing},
  pages={21445--21482},
  year={2025}
}

@inproceedings{singh2023nlms,
  title={NLMs: Augmenting negation in language models},
  author={Singh, Rituraj and Kumar, Rahul and Sridhar, Vivek},
  booktitle={Findings of the Association for Computational Linguistics: EMNLP 2023},
  pages={13104--13116},
  year={2023}
}

@article{jurafsky2014speech,
  title={Speech and Language Processing/Daniel Jurafsky \& James H},
  author={Jurafsky, Daniel},
  journal={New International ed. Harlow: Pearson Education, c2014., Harlow},
  year={2014}
}

@article{carter2024discourse,
  title={Discourse coherence modulates use of predictive processing during sentence comprehension},
  author={Carter, Georgia-Ann and Hoffman, Paul},
  journal={Cognition},
  volume={242},
  pages={105637},
  year={2024},
  publisher={Elsevier}
}

@article{lu2025principle,
  title={The principle of anticipation in language use},
  author={Lu, Fangzhe and Ursini, Francesco-Alessio and Zhu, Bin and Yuan, Chenjie and Zeng, Jun},
  journal={Humanities and Social Sciences Communications},
  volume={12},
  number={1},
  pages={1--17},
  year={2025},
  publisher={Palgrave}
}

@article{pedregosa2011scikit,
  title={Scikit-learn: Machine learning in Python},
  author={Pedregosa, Fabian and Varoquaux, Ga{\"e}l and Gramfort, Alexandre and Michel, Vincent and Thirion, Bertrand and Grisel, Olivier and Blondel, Mathieu and Prettenhofer, Peter and Weiss, Ron and Dubourg, Vincent and others},
  journal={the Journal of machine Learning research},
  volume={12},
  pages={2825--2830},
  year={2011},
  publisher={JMLR. org}
}

@article{grattafiori2024llama,
  title={The llama 3 herd of models},
  author={Grattafiori, Aaron and Dubey, Abhimanyu and Jauhri, Abhinav and Pandey, Abhinav and Kadian, Abhishek and Al-Dahle, Ahmad and Letman, Aiesha and Mathur, Akhil and Schelten, Alan and Vaughan, Alex and others},
  journal={arXiv preprint arXiv:2407.21783},
  year={2024}
}

@article{morante2011annotation,
  title={Annotation of negation cues and their scope: Guidelines v1},
  author={Morante, Roser and Schrauwen, Sarah and Daelemans, Walter},
  journal={Computational linguistics and psycholinguistics technical report series, CTRS-003},
  pages={1--42},
  year={2011}
}

@inproceedings{hossain-etal-2020-non,
    title = "It{'}s not a Non-Issue: Negation as a Source of Error in Machine Translation",
    author = "Hossain, Md Mosharaf  and
      Anastasopoulos, Antonios  and
      Blanco, Eduardo  and
      Palmer, Alexis",
    editor = "Cohn, Trevor  and
      He, Yulan  and
      Liu, Yang",
    booktitle = "Findings of the Association for Computational Linguistics: EMNLP 2020",
    month = nov,
    year = "2020",
    address = "Online",
    publisher = "Association for Computational Linguistics",
    url = "https://aclanthology.org/2020.findings-emnlp.345/",
    doi = "10.18653/v1/2020.findings-emnlp.345",
    pages = "3869--3885"
}

@article{ettinger-2020-bert,
    title = "What {BERT} Is Not: Lessons from a New Suite of Psycholinguistic Diagnostics for Language Models",
    author = "Ettinger, Allyson",
    editor = "Johnson, Mark  and
      Roark, Brian  and
      Nenkova, Ani",
    journal = "Transactions of the Association for Computational Linguistics",
    volume = "8",
    year = "2020",
    address = "Cambridge, MA",
    publisher = "MIT Press",
    url = "https://aclanthology.org/2020.tacl-1.3/",
    doi = "10.1162/tacl_a_00298",
    pages = "34--48"
}

@inproceedings{szarvas-etal-2008-bioscope,
    title = "The {B}io{S}cope corpus: annotation for negation, uncertainty and their scope in biomedical texts",
    author = {Szarvas, Gy{\"o}rgy  and
      Vincze, Veronika  and
      Farkas, Rich{\'a}rd  and
      Csirik, J{\'a}nos},
    editor = "Demner-Fushman, Dina  and
      Ananiadou, Sophia  and
      Cohen, Kevin Bretonnel  and
      Pestian, John  and
      Tsujii, Jun{'}ichi  and
      Webber, Bonnie",
    booktitle = "Proceedings of the Workshop on Current Trends in Biomedical Natural Language Processing",
    month = jun,
    year = "2008",
    address = "Columbus, Ohio",
    publisher = "Association for Computational Linguistics",
    url = "https://aclanthology.org/W08-0606/",
    pages = "38--45"
}

@InCollection{sep-negation,
	author       =	{Horn, Laurence R. and Wansing, Heinrich},
	title        =	{{Negation}},
	booktitle    =	{The {Stanford} Encyclopedia of Philosophy},
	editor       =	{Edward N. Zalta and Uri Nodelman},
	howpublished =	{\url{https://plato.stanford.edu/archives/spr2025/entries/negation/}},
	year         =	{2025},
	edition      =	{{S}pring 2025},
	publisher    =	{Metaphysics Research Lab, Stanford University}
}

@book{english.grammar.2002,
  author = {Huddleston, Rodney D. and Pullum, Geoffrey K.},
  month = {April},
  publisher = {Cambridge University Press},
  title = {The Cambridge Grammar of the English Language},
  year = 2002
}

@inproceedings{morante-daelemans-2012-conandoyle,
    title = "{C}onan{D}oyle-neg: Annotation of negation cues and their scope in Conan Doyle stories",
    author = "Morante, Roser  and
      Daelemans, Walter",
    editor = "Calzolari, Nicoletta  and
      Choukri, Khalid  and
      Declerck, Thierry  and
      Do{\u{g}}an, Mehmet U{\u{g}}ur  and
      Maegaard, Bente  and
      Mariani, Joseph  and
      Moreno, Asuncion  and
      Odijk, Jan  and
      Piperidis, Stelios",
    booktitle = "Proceedings of the Eighth International Conference on Language Resources and Evaluation ({LREC}'12)",
    month = may,
    year = "2012",
    address = "Istanbul, Turkey",
    publisher = "European Language Resources Association (ELRA)",
    url = "https://aclanthology.org/L12-1077/",
    pages = "1563--1568"
}

@inproceedings{lapponi-etal-2012-uio,
    title = "{U}i{O} 2: Sequence-labeling Negation Using Dependency Features",
    author = "Lapponi, Emanuele  and
      Velldal, Erik  and
      {\O}vrelid, Lilja  and
      Read, Jonathon",
    editor = "Agirre, Eneko  and
      Bos, Johan  and
      Diab, Mona  and
      Manandhar, Suresh  and
      Marton, Yuval  and
      Yuret, Deniz",
    booktitle = "*{SEM} 2012: The First Joint Conference on Lexical and Computational Semantics {--} Volume 1: Proceedings of the main conference and the shared task, and Volume 2: Proceedings of the Sixth International Workshop on Semantic Evaluation ({S}em{E}val 2012)",
    month = "7-8 " # jun,
    year = "2012",
    address = "Montr{\'e}al, Canada",
    publisher = "Association for Computational Linguistics",
    url = "https://aclanthology.org/S12-1042/",
    pages = "319--327"
}

@inproceedings{hossain-etal-2020-analysis,
    title = "An Analysis of Natural Language Inference Benchmarks through the Lens of Negation",
    author = "Hossain, Md Mosharaf  and
      Kovatchev, Venelin  and
      Dutta, Pranoy  and
      Kao, Tiffany  and
      Wei, Elizabeth  and
      Blanco, Eduardo",
    editor = "Webber, Bonnie  and
      Cohn, Trevor  and
      He, Yulan  and
      Liu, Yang",
    booktitle = "Proceedings of the 2020 Conference on Empirical Methods in Natural Language Processing (EMNLP)",
    month = nov,
    year = "2020",
    address = "Online",
    publisher = "Association for Computational Linguistics",
    url = "https://aclanthology.org/2020.emnlp-main.732/",
    doi = "10.18653/v1/2020.emnlp-main.732",
    pages = "9106--9118"
}

@inproceedings{ravichander-etal-2022-condaqa,
    title = "{CONDAQA}: A Contrastive Reading Comprehension Dataset for Reasoning about Negation",
    author = "Ravichander, Abhilasha  and
      Gardner, Matt  and
      Marasovic, Ana",
    editor = "Goldberg, Yoav  and
      Kozareva, Zornitsa  and
      Zhang, Yue",
    booktitle = "Proceedings of the 2022 Conference on Empirical Methods in Natural Language Processing",
    month = dec,
    year = "2022",
    address = "Abu Dhabi, United Arab Emirates",
    publisher = "Association for Computational Linguistics",
    url = "https://aclanthology.org/2022.emnlp-main.598/",
    doi = "10.18653/v1/2022.emnlp-main.598",
    pages = "8729--8755"
}

@inproceedings{rezaei-blanco-2024-paraphrasing,
    title = "Paraphrasing in Affirmative Terms Improves Negation Understanding",
    author = "Rezaei, MohammadHossein  and
      Blanco, Eduardo",
    editor = "Ku, Lun-Wei  and
      Martins, Andre  and
      Srikumar, Vivek",
    booktitle = "Proceedings of the 62nd Annual Meeting of the Association for Computational Linguistics (Volume 2: Short Papers)",
    month = aug,
    year = "2024",
    address = "Bangkok, Thailand",
    publisher = "Association for Computational Linguistics",
    url = "https://aclanthology.org/2024.acl-short.55/",
    doi = "10.18653/v1/2024.acl-short.55",
    pages = "602--615"
}

@inproceedings{rezaei-blanco-2025-making,
    title = "Making Language Models Robust Against Negation",
    author = "Rezaei, MohammadHossein  and
      Blanco, Eduardo",
    editor = "Chiruzzo, Luis  and
      Ritter, Alan  and
      Wang, Lu",
    booktitle = "Proceedings of the 2025 Conference of the Nations of the Americas Chapter of the Association for Computational Linguistics: Human Language Technologies (Volume 1: Long Papers)",
    month = apr,
    year = "2025",
    address = "Albuquerque, New Mexico",
    publisher = "Association for Computational Linguistics",
    url = "https://aclanthology.org/2025.naacl-long.413/",
    doi = "10.18653/v1/2025.naacl-long.413",
    pages = "8123--8142",
    ISBN = "979-8-89176-189-6"
}

@inproceedings{hosseini-etal-2021-understanding,
    title = "Understanding by Understanding Not: Modeling Negation in Language Models",
    author = "Hosseini, Arian  and
      Reddy, Siva  and
      Bahdanau, Dzmitry  and
      Hjelm, R Devon  and
      Sordoni, Alessandro  and
      Courville, Aaron",
    editor = "Toutanova, Kristina  and
      Rumshisky, Anna  and
      Zettlemoyer, Luke  and
      Hakkani-Tur, Dilek  and
      Beltagy, Iz  and
      Bethard, Steven  and
      Cotterell, Ryan  and
      Chakraborty, Tanmoy  and
      Zhou, Yichao",
    booktitle = "Proceedings of the 2021 Conference of the North American Chapter of the Association for Computational Linguistics: Human Language Technologies",
    month = jun,
    year = "2021",
    address = "Online",
    publisher = "Association for Computational Linguistics",
    url = "https://aclanthology.org/2021.naacl-main.102/",
    doi = "10.18653/v1/2021.naacl-main.102",
    pages = "1301--1312"
}

@article{mcnemar1947note,
  title={Note on the sampling error of the difference between correlated proportions or percentages},
  author={McNemar, Quinn},
  journal={Psychometrika},
  volume={12},
  number={2},
  pages={153--157},
  year={1947},
  publisher={Springer-Verlag}
}

@inproceedings{hossain-blanco-2022-leveraging,
  title = {Leveraging Affirmative Interpretations from Negation Improves Natural Language Understanding},
  author = {Hossain, Md Mosharaf and Blanco, Eduardo},
  booktitle = {Proceedings of the 2022 Conference on Empirical Methods in Natural Language Processing},
  month = dec,
  year = {2022},
  address = {Abu Dhabi, United Arab Emirates},
  publisher = {Association for Computational Linguistics},
  url = {https://aclanthology.org/2022.emnlp-main.393},
  pages = {5833--5847},
  month_numeric = {12}
}

@inproceedings{zhang-etal-2023-beyond,
    title = "Beyond Positive Scaling: How Negation Impacts Scaling Trends of Language Models",
    author = "Zhang, Yuhui  and
      Yasunaga, Michihiro  and
      Zhou, Zhengping  and
      HaoChen, Jeff Z.  and
      Zou, James  and
      Liang, Percy  and
      Yeung, Serena",
    editor = "Rogers, Anna  and
      Boyd-Graber, Jordan  and
      Okazaki, Naoaki",
    booktitle = "Findings of the Association for Computational Linguistics: ACL 2023",
    month = jul,
    year = "2023",
    address = "Toronto, Canada",
    publisher = "Association for Computational Linguistics",
    url = "https://aclanthology.org/2023.findings-acl.472/",
    doi = "10.18653/v1/2023.findings-acl.472",
    pages = "7479--7498"
}

@inproceedings{weller-etal-2024-nevir,
    title = "{N}ev{IR}: Negation in Neural Information Retrieval",
    author = "Weller, Orion  and
      Lawrie, Dawn  and
      Van Durme, Benjamin",
    editor = "Graham, Yvette  and
      Purver, Matthew",
    booktitle = "Proceedings of the 18th Conference of the European Chapter of the Association for Computational Linguistics (Volume 1: Long Papers)",
    month = mar,
    year = "2024",
    address = "St. Julian{'}s, Malta",
    publisher = "Association for Computational Linguistics",
    url = "https://aclanthology.org/2024.eacl-long.139/",
    pages = "2274--2287"
}

@inproceedings{she-etal-2023-scone,
    title = "{S}co{N}e: Benchmarking Negation Reasoning in Language Models With Fine-Tuning and In-Context Learning",
    author = "She, Jingyuan S.  and
      Potts, Christopher  and
      Bowman, Samuel R.  and
      Geiger, Atticus",
    editor = "Rogers, Anna  and
      Boyd-Graber, Jordan  and
      Okazaki, Naoaki",
    booktitle = "Proceedings of the 61st Annual Meeting of the Association for Computational Linguistics (Volume 2: Short Papers)",
    month = jul,
    year = "2023",
    address = "Toronto, Canada",
    publisher = "Association for Computational Linguistics",
    url = "https://aclanthology.org/2023.acl-short.154/",
    doi = "10.18653/v1/2023.acl-short.154",
    pages = "1803--1821"
}

@misc{wikidump,
    author = "Wikimedia-Foundation",
    title  = "Wikimedia Downloads",
    url    = "https://dumps.wikimedia.org",
    year   = "2023"
}

@inproceedings{van-son-etal-2016-building,
    title = "Building a Dictionary of Affixal Negations",
    author = "van Son, Chantal  and
      van Miltenburg, Emiel  and
      Morante, Roser",
    editor = "Blanco, Eduardo  and
      Morante, Roser  and
      Saur{\'i}, Roser",
    booktitle = "Proceedings of the Workshop on Extra-Propositional Aspects of Meaning in Computational Linguistics ({E}x{P}ro{M})",
    month = dec,
    year = "2016",
    address = "Osaka, Japan",
    publisher = "The COLING 2016 Organizing Committee",
    url = "https://aclanthology.org/W16-5007/",
    pages = "49--56"
}

@article{joshi2012affixal,
  title={Affixal negation--direct, indirect and their subtypes},
  author={Joshi, Shrikant},
  journal={Syntaxe \& s{\'e}mantique},
  volume={13},
  number={1},
  pages={49--63},
  year={2012},
  publisher={Presses universitaires de Caen}
}

@article{dettmers2023qlora,
  title={Qlora: Efficient finetuning of quantized llms},
  author={Dettmers, Tim and Pagnoni, Artidoro and Holtzman, Ari and Zettlemoyer, Luke},
  journal={Advances in neural information processing systems},
  volume={36},
  pages={10088--10115},
  year={2023}
}

@misc{yang2024qwen2technicalreport,
      title={Qwen2 Technical Report}, 
      author={An Yang and Baosong Yang and Binyuan Hui and Bo Zheng and Bowen Yu and Chang Zhou and Chengpeng Li and Chengyuan Li and Dayiheng Liu and Fei Huang and Guanting Dong and Haoran Wei and Huan Lin and Jialong Tang and Jialin Wang and Jian Yang and Jianhong Tu and Jianwei Zhang and Jianxin Ma and Jianxin Yang and Jin Xu and Jingren Zhou and Jinze Bai and Jinzheng He and Junyang Lin and Kai Dang and Keming Lu and Keqin Chen and Kexin Yang and Mei Li and Mingfeng Xue and Na Ni and Pei Zhang and Peng Wang and Ru Peng and Rui Men and Ruize Gao and Runji Lin and Shijie Wang and Shuai Bai and Sinan Tan and Tianhang Zhu and Tianhao Li and Tianyu Liu and Wenbin Ge and Xiaodong Deng and Xiaohuan Zhou and Xingzhang Ren and Xinyu Zhang and Xipin Wei and Xuancheng Ren and Xuejing Liu and Yang Fan and Yang Yao and Yichang Zhang and Yu Wan and Yunfei Chu and Yuqiong Liu and Zeyu Cui and Zhenru Zhang and Zhifang Guo and Zhihao Fan},
      year={2024},
      eprint={2407.10671},
      archivePrefix={arXiv},
      primaryClass={cs.CL},
      url={https://arxiv.org/abs/2407.10671}, 
}

@article{liu2019roberta,
  title={Roberta: A robustly optimized bert pretraining approach},
  author={Liu, Yinhan and Ott, Myle and Goyal, Naman and Du, Jingfei and Joshi, Mandar and Chen, Danqi and Levy, Omer and Lewis, Mike and Zettlemoyer, Luke and Stoyanov, Veselin},
  journal={arXiv preprint arXiv:1907.11692},
  year={2019}
}

@inproceedings{devlin-etal-2019-bert,
    title = "{BERT}: Pre-training of Deep Bidirectional Transformers for Language Understanding",
    author = "Devlin, Jacob  and
      Chang, Ming-Wei  and
      Lee, Kenton  and
      Toutanova, Kristina",
    editor = "Burstein, Jill  and
      Doran, Christy  and
      Solorio, Thamar",
    booktitle = "Proceedings of the 2019 Conference of the North {A}merican Chapter of the Association for Computational Linguistics: Human Language Technologies, Volume 1 (Long and Short Papers)",
    month = jun,
    year = "2019",
    address = "Minneapolis, Minnesota",
    publisher = "Association for Computational Linguistics",
    url = "https://aclanthology.org/N19-1423/",
    doi = "10.18653/v1/N19-1423",
    pages = "4171--4186"
}

@inproceedings{wei-etal-2023-inverse,
    title = "Inverse Scaling Can Become {U}-Shaped",
    author = "Wei, Jason  and
      Kim, Najoung  and
      Tay, Yi  and
      Le, Quoc",
    editor = "Bouamor, Houda  and
      Pino, Juan  and
      Bali, Kalika",
    booktitle = "Proceedings of the 2023 Conference on Empirical Methods in Natural Language Processing",
    month = dec,
    year = "2023",
    address = "Singapore",
    publisher = "Association for Computational Linguistics",
    url = "https://aclanthology.org/2023.emnlp-main.963/",
    doi = "10.18653/v1/2023.emnlp-main.963",
    pages = "15580--15591"
}

@misc{mckenzie2024inversescalingbiggerisnt,
      title={Inverse Scaling: When Bigger Isn't Better}, 
      author={Ian R. McKenzie and Alexander Lyzhov and Michael Pieler and Alicia Parrish and Aaron Mueller and Ameya Prabhu and Euan McLean and Aaron Kirtland and Alexis Ross and Alisa Liu and Andrew Gritsevskiy and Daniel Wurgaft and Derik Kauffman and Gabriel Recchia and Jiacheng Liu and Joe Cavanagh and Max Weiss and Sicong Huang and The Floating Droid and Tom Tseng and Tomasz Korbak and Xudong Shen and Yuhui Zhang and Zhengping Zhou and Najoung Kim and Samuel R. Bowman and Ethan Perez},
      year={2024},
      eprint={2306.09479},
      archivePrefix={arXiv},
      primaryClass={cs.CL},
      url={https://arxiv.org/abs/2306.09479}, 
}

@inproceedings{geiger-etal-2020-neural,
    title = "Neural Natural Language Inference Models Partially Embed Theories of Lexical Entailment and Negation",
    author = "Geiger, Atticus  and
      Richardson, Kyle  and
      Potts, Christopher",
    editor = "Alishahi, Afra  and
      Belinkov, Yonatan  and
      Chrupa{\l}a, Grzegorz  and
      Hupkes, Dieuwke  and
      Pinter, Yuval  and
      Sajjad, Hassan",
    booktitle = "Proceedings of the Third BlackboxNLP Workshop on Analyzing and Interpreting Neural Networks for NLP",
    month = nov,
    year = "2020",
    address = "Online",
    publisher = "Association for Computational Linguistics",
    url = "https://aclanthology.org/2020.blackboxnlp-1.16/",
    doi = "10.18653/v1/2020.blackboxnlp-1.16",
    pages = "163--173"
}

@inproceedings{bowman-etal-2015-large,
    title = "A large annotated corpus for learning natural language inference",
    author = "Bowman, Samuel R.  and
      Angeli, Gabor  and
      Potts, Christopher  and
      Manning, Christopher D.",
    editor = "M{\`a}rquez, Llu{\'i}s  and
      Callison-Burch, Chris  and
      Su, Jian",
    booktitle = "Proceedings of the 2015 Conference on Empirical Methods in Natural Language Processing",
    month = sep,
    year = "2015",
    address = "Lisbon, Portugal",
    publisher = "Association for Computational Linguistics",
    url = "https://aclanthology.org/D15-1075/",
    doi = "10.18653/v1/D15-1075",
    pages = "632--642"
}

@book{horn1989natural,
  title     = {A Natural History of Negation},
  author    = {Horn, Laurence R.},
  year      = {1989},
  publisher = {University of Chicago Press},
  address   = {Chicago}
}

@article{brown2020language,
  title={Language models are few-shot learners},
  author={Brown, Tom and Mann, Benjamin and Ryder, Nick and Subbiah, Melanie and Kaplan, Jared D and Dhariwal, Prafulla and Neelakantan, Arvind and Shyam, Pranav and Sastry, Girish and Askell, Amanda and others},
  journal={Advances in neural information processing systems},
  volume={33},
  pages={1877--1901},
  year={2020}
}

@inproceedings{reimers-gurevych-2019-sentence,
    title = "Sentence-{BERT}: Sentence Embeddings using {S}iamese {BERT}-Networks",
    author = "Reimers, Nils  and
      Gurevych, Iryna",
    editor = "Inui, Kentaro  and
      Jiang, Jing  and
      Ng, Vincent  and
      Wan, Xiaojun",
    booktitle = "Proceedings of the 2019 Conference on Empirical Methods in Natural Language Processing and the 9th International Joint Conference on Natural Language Processing (EMNLP-IJCNLP)",
    month = nov,
    year = "2019",
    address = "Hong Kong, China",
    publisher = "Association for Computational Linguistics",
    url = "https://aclanthology.org/D19-1410/",
    doi = "10.18653/v1/D19-1410",
    pages = "3982--3992"
}

@article{vaswani2017attention,
  title={Attention is all you need},
  author={Vaswani, Ashish and Shazeer, Noam and Parmar, Niki and Uszkoreit, Jakob and Jones, Llion and Gomez, Aidan N and Kaiser, {\L}ukasz and Polosukhin, Illia},
  journal={Advances in neural information processing systems},
  volume={30},
  year={2017}
}

@article{wang2022benchmarking,
  title={Benchmarking generalization via in-context instructions on 1,600+ language tasks},
  author={Wang, Yizhong and Mishra, Swaroop and Alipoormolabashi, Pegah and Kordi, Yeganeh and Mirzaei, Amirreza and Arunkumar, Anjana and Ashok, Arjun and Dhanasekaran, Arut Selvan and Naik, Atharva and Stap, David and others},
  journal={arXiv preprint arXiv:2204.07705},
  volume={2},
  pages={2},
  year={2022}
}

@article{burns2022discovering,
  title={Discovering latent knowledge in language models without supervision},
  author={Burns, Collin and Ye, Haotian and Klein, Dan and Steinhardt, Jacob},
  journal={arXiv preprint arXiv:2212.03827},
  year={2022}
}

\clearpage
\appendix
% %%%%%%%%%%%%%%%%%%%%%%%%%%%%%%%%%%%%%%%%%%%%%%%%%%%%%%%%%%%%%%
% Appendix
% %%%%%%%%%%%%%%%%%%%%%%%%%%%%%%%%%%%%%%%%%%%%%%%%%%%%%%%%%%%%%%

% %%%%%%%%%%%%%%%%%%%%%%%%%%%%%%%%%%%%%%%%%%%%%%%%%%%%%%%%%%%%%%
% NSPP validation
% %%%%%%%%%%%%%%%%%%%%%%%%%%%%%%%%%%%%%%%%%%%%%%%%%%%%%%%%%%%%%%
\section{Validating the Learnability of NSPP}
\label{app:nspp_validate}
In the NSPP task, we construct sentence-pair samples $(S_1, S_2)$. Samples are labeled as NEG if $S_2$ contains negation, and AFF otherwise. During training, however, only $S_1$ is provided to the model, which is asked to predict whether the following sentence $S_2$ contains negation.

To examine whether this task is genuinely learnable, we design two simple yet intuitive linear probing experiments. 
Although previous work has shown that NSPP improves the robustness of language models to negation, 
our goal here is to directly investigate whether the latent semantic signal required by NSPP can be learned from $S_1$ alone. 
Intuitively, if even a simple linear model can capture such signals and benefit from them, then the substantially more expressive language models used in our work should also be capable of learning them.

To provide a more comprehensive analysis, we study the task from two fine-grained perspectives. 
The first is \textbf{polarity prediction}, where the model is given only $S_1$ and asked to predict whether the following sentence $S_2$ contains negation. 
The second, and more challenging, setting is \textbf{negation type prediction}. Here, the model is given only the $S_1$ sentences from NEG samples and must predict the specific negation type appearing in $S_2$. In the following sections, we describe these two experiments and their results in detail.

\subsection{Polarity Prediction}
For the polarity prediction task, we randomly sample 5K AFF and 5K NEG samples from the NegCue training set. 
Using the Sentence-Transformers framework, each $S_1$ sentence is converted into a 384-dimensional embedding. 
The data are split into training and test sets with an 8:2 ratio. 
We then train a simple logistic regression classifier from \textit{sklearn}~\cite{pedregosa2011scikit} on the frozen $S_1$ embeddings to predict whether the following sentence $S_2$ contains negation (AFF vs.\ NEG). 
To improve robustness, we repeat the experiments with five random seeds and report both accuracy and Macro F1 scores. 
We additionally analyze the resulting confusion matrices.
\begin{table}[t]
\centering
\small
\begin{tabular}{lccc}
\toprule
\textbf{Task} & \textbf{Random} & \textbf{Acc.} & \textbf{Macro F1} \\
\midrule
Polarity & 50.0 & 57.7 $\pm$ 0.45 & 57.7 $\pm$ 0.44 \\
Negation Type & 33.3 & 39.6 $\pm$ 0.24 & 39.6 $\pm$ 0.24 \\
\bottomrule
\end{tabular}
\caption{Linear probing results using only frozen $S_1$ embeddings to predict the polarity or negation type of the following sentence $S_2$.}
\label{t:linear_probe_results}
\end{table}

Table~\ref{t:linear_probe_results} shows that, compared to the random baseline (50\%), a simple linear model trained on only 10K samples achieves nearly 58\% accuracy. 
This result is substantially above random guessing, suggesting that the model is able to learn meaningful patterns from $S_1$ alone. 
Figure~\ref{fig:confusion_matrix} presents the confusion matrices for our linear probing experiments. 
In the polarity prediction setting (left), the higher diagonal values show that the model learns to distinguish AFF and NEG samples rather than collapsing to majority-class predictions. 
This suggests that the $S_1$ embeddings contain meaningful signals for inferring whether the following sentence $S_2$ contains negation.

\subsection{Negation Type Prediction}
We further investigate whether, given only $S_1$, the model can distinguish the negation type appearing in $S_2$. As discussed in Section~2, prior linguistic studies have shown that coherent adjacent sentences often exhibit semantic and logical dependencies, and such discourse structure may even implicitly constrain how negation is expressed in the following sentence.

\begin{figure}[t]
    \centering
    \includegraphics[width=\linewidth]{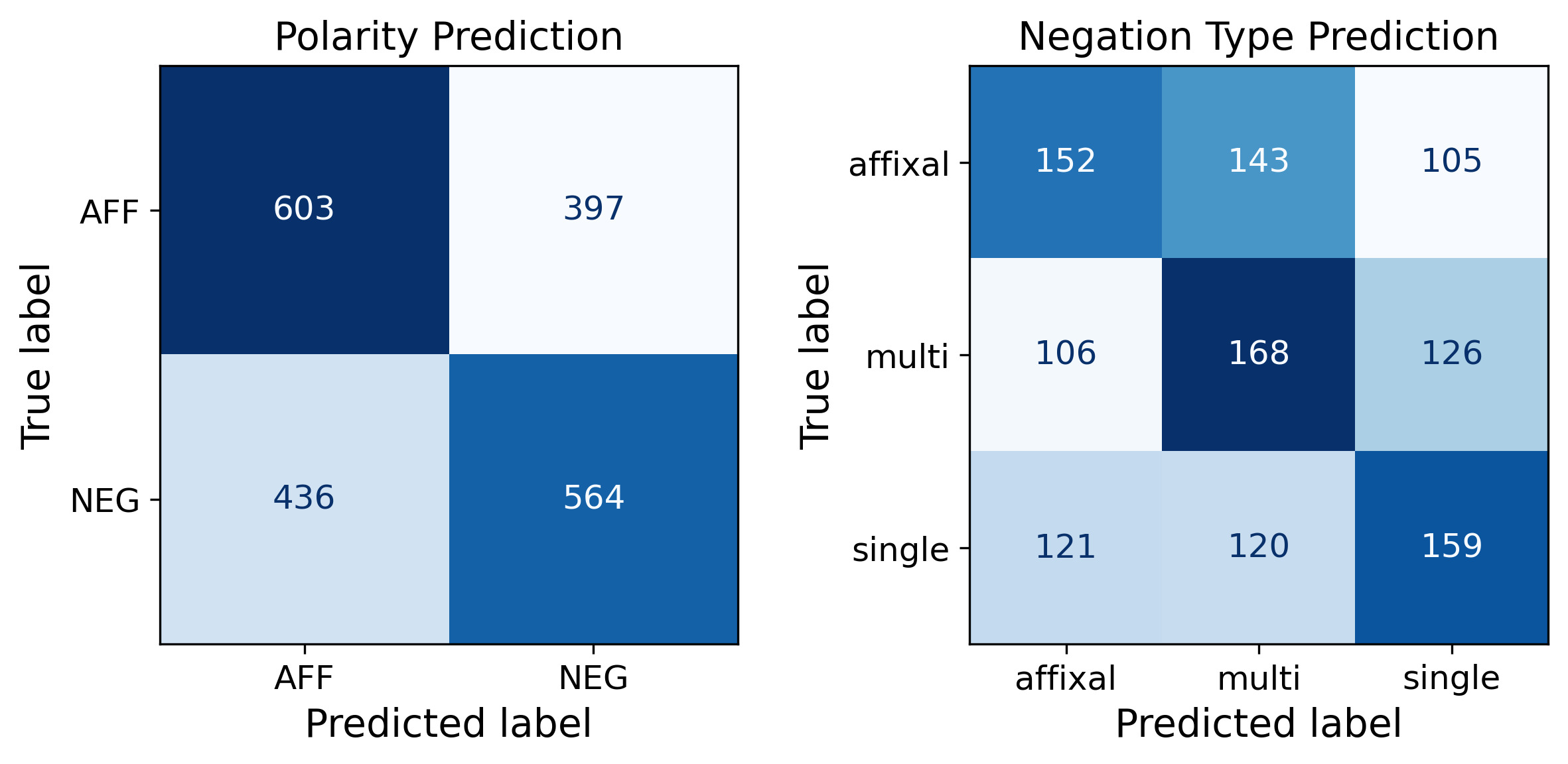}
    \caption{Confusion matrices for the two linear probing experiments. Left: polarity prediction (AFF vs.\ NEG) using only $S_1$ embeddings. Right: negation type prediction (affixal, multi-word, and single-word negation) using only $S_1$ embeddings from NEG samples.}
    \label{fig:confusion_matrix}
\end{figure}

To study this, we randomly sample 6K NEG samples from NegCue, balanced across the three negation types: single-word, multi-word, and affixal negation. 
The data are split into training and test sets with an 8:2 ratio, and the $S_1$ sentences are converted into embeddings for a three-way classification task. 
We train the same linear model for 5K iterations and report the results in Table~\ref{t:linear_probe_results}.

Despite the increased difficulty of this task, the simple linear model still achieves nearly 40\% accuracy, substantially above the random baseline (33.3\%). 
Moreover, the confusion matrix in Figure~\ref{fig:confusion_matrix} (right) shows that the model does not rely on random guessing, but instead learns meaningful distinctions between negation types. 
Interestingly, affixal negation appears to be the most difficult type to predict correctly, which may also explain why further pre-training on it yields larger improvements in our language model experiments. 
This observation is consistent with our earlier findings that affixal negation remains comparatively more challenging for current language models.

Overall, these results suggest that $S_1$ contains learnable semantic signals associated not only with whether negation appears in $S_2$, but also with the type of negation that is likely to occur. We therefore have strong reason to expect that more powerful language models can learn negation-related knowledge through the NSPP task.

% %%%%%%%%%%%%%%%%%%%%%%%%%%%%%%%%%%%%%%%%%%%%%%%%%%%%%%%%%%%%%%
% NegCue
% %%%%%%%%%%%%%%%%%%%%%%%%%%%%%%%%%%%%%%%%%%%%%%%%%%%%%%%%%%%%%%
\section{NegCue Corpus}
As discussed in Section~\ref{subs:negcue}, constructing a corpus that comprehensively covers diverse negation types is far from straightforward. 
In this section, we provide additional details about the construction of NegCue. 
We also conduct a two-part manual verification process. 
First, we verify the categorization of all extracted unique negation cues to ensure that cues are correctly assigned to their corresponding negation types. 
Second, we evaluate the quality of the constructed NegCue sentence-pair samples.
\label{app:negcue_details}
\subsection{Construction Details}
Although the CondaQA dataset provides an \textit{original cue} field, we do not directly rely on it for cue identification, 
as it includes tokens such as \textit{could} and \textit{loss} that do not explicitly express negation.
As a result, we construct a refined cue list derived from the original CondaQA cues.

Using this cue list, we traverse the full Wikipedia corpus and extract sentence pairs that meet our criteria. 
Among the three cue types, multi-word cues yield the smallest number of samples, with approximately 309,857 samples. 
To construct a roughly class-balanced corpus, we pair these negative (NEG) samples with an equal number of 
affirmative (AFF) sentence pairs that do not contain negation, resulting in a multi-word corpus of 619,714 samples. 
In addition, to ensure that AFF and NEG samples remain as comparable as possible in domain and semantic content, we extract matched affirmative and negative sentence pairs from the same Wikipedia articles whenever possible.
We then randomly split each corpus into training, validation, and test sets using a 90:5:5 ratio, 
ensuring that all splits contain the corresponding negation cues and exhibit similar cue distributions.

Following the same procedure, we construct two additional corpora for the single-word and affixal cue types. 
Finally, we separately merge the corresponding training, validation, and test splits of the three cue-specific corpora to construct the \textit{All Cues} corpus. 
The \textit{Single-word}, \textit{Multi-word}, \textit{Affixal}, and \textit{All Cues} corpora together constitute NegCue. 
We provide several real examples in Figure~\ref{fig:negcue_examples}.

\begin{figure}[t]
\small

\begin{tcolorbox}[
    colback=gray!5,
    colframe=gray!70,
    boxrule=0.5pt,
    arc=2pt,
    left=6pt,
    right=6pt,
    top=4pt,
    bottom=4pt
]

\textbf{Example 1}

\vspace{0.3em}

\textbf{S$_1$:} The council criticized his conduct but agreed to vote him the £100 salary for his year's services as mayor.

\vspace{0.3em}

\textbf{S$_2$:} Ford \underline{refused} to accept the salary with censure and brought the affair to public attention.

\vspace{0.6em}

\noindent\rule{\textwidth}{0.4pt}

\vspace{0.6em}

\textbf{Example 2}

\vspace{0.3em}

\textbf{S$_1$:} Consequently, there is a greater clarity and cohesion to the Director's Cut.

\vspace{0.3em}

\textbf{S$_2$:} We are \underline{not} going to make both versions available.

\end{tcolorbox}

\caption{
Real examples from the NegCue corpus.
}

\label{fig:negcue_examples}

\end{figure}

\subsection{Manual Inspection}
In Section~\ref{subs:negcue}, we discuss how the complexity and diversity of negation cues make accurate cue identification and categorization challenging.
Although we employ carefully designed rule-based detection methods together with a cue dictionary for automatic classification, 
we additionally conduct a manual verification of all extracted cues to ensure that each cue is correctly assigned to its corresponding negation type.
Our target cue list contains 214 negation cues in total. 
After independent manual verification by two annotators, all cues were confirmed to be correctly categorized. 
Among them, 167 are affixal negation cues, 36 are single-word negation cues, and 11 are multi-word negation cues.

In addition, to evaluate the quality of the sentence-pair samples extracted from Wikipedia, we randomly sample 500 samples from the NegCue training set for manual inspection. 
Among them, 262 are labeled as AFF and 238 as NEG, indicating that the corpus is approximately class-balanced. 
We then design four evaluation criteria to comprehensively assess sample quality:
\begin{itemize}[itemsep=1pt, nosep]
    \item Whether the sample labels are correct, i.e., whether each sample truly satisfies the definition of an AFF or NEG samples.
    \item Whether both sentences in the $(S_1, S_2)$ pair are complete and valid sentences rather than incorrectly truncated fragments.
    \item For NEG samples, whether the negation cue type is correctly categorized as single-word, multi-word, or affixal negation.
    \item Whether the samples contain negation cues outside our predefined cue list.
\end{itemize}

Overall, NegCue demonstrates high-quality sample construction across all evaluation criteria. 
AFF/NEG label correctness reaches 99.8\%, while negation cue type classification achieves 100\% accuracy. 
Sentence validity is also high at 98.4\%, and 
99.4\% of the samples do not contain unseen negation cues outside our predefined cue list.

During manual inspection, we identified only one AFF sample whose $S_2$ sentence unexpectedly contained a negation cue not included in our predefined cue inventory, causing it to be incorrectly labeled as affirmative.
In addition, eight samples contained incomplete sentences caused by sentence segmentation errors, 
typically due to periods appearing inside entity names or abbreviations such as ``Museo L. Pigorini'' or ``N.Y.D.''. 
Finally, only three samples contained negation cues outside our 214-cue list. 
Taken together, these results indicate that NegCue maintains high annotation quality and reliable negation cue categorization despite the complexity and diversity of negation phenomena.

\begin{table*}[t!]
  \centering
  \small
  \begin{tabular}{l r@{ }l r@{ }l r@{ }l r@{ }l rrrr}
\toprule
&  \multicolumn{8}{c}{\textbf{Accuracy}}  & \multicolumn{4}{c}{\textbf{Group Consistency}} \\
 \cmidrule(lr){2-9} \cmidrule(lr){10-13}

& All &($\Delta$\%) & Affixal  & & Single  & & Multi & & All & Par. & Sco. & Aff.  \\ \midrule

Qwen2-0.5B  
& 49.6 & & 48.4 & & 50.9 & & 49.6 & & 3.8 & 47.0 & 18.4 & 15.2 \\
~~~~further pre-trained: \\
~~~~~~~affixal cues    
& 49.9 &\scriptsize($+$0.7) & 49.6 & & \textbf{50.7} & & 46.2 & & \textbf{5.1} & 45.1 & 17.4 & \textbf{17.4} \\
~~~~~~~single-word cues 
& 48.4 &\scriptsize($-$2.5) & 48.1 & & 49.0 & & 44.2 & & 4.1 & 44.9 & 16.1 & 15.6 \\
~~~~~~~multi-word cues  
& \textbf{50.0} &\scriptsize($+$0.8) & \textbf{49.8} & & 50.2 & & \textbf{48.9} & & 4.7 & \textbf{46.1} & \textbf{17.6} & \textbf{17.4} \\
~~~~~~~All cues    
& 48.9 &\scriptsize($-$1.4) & 48.8 & & 49.3 & & 45.4 & & 4.3 & 45.2 & 17.0 & 16.6 \\

\midrule
\midrule

& $Q_1$ & & $Q_2$ & & $Q_3$ & & $Q_4$ & & $Q_1$ & $Q_2$ & $Q_3$ & $Q_4$  \\
\cmidrule(lr){2-9} \cmidrule(lr){10-13}

Qwen2-0.5B 
& 50.2 && 50.0 && 46.1 && 45.7 && 4.9  & 2.1  & 2.6  & 0.0 \\
~~~+ NSPP, multi-word 
& 50.5 &\scriptsize$(+0.6)$ & 48.5 &\scriptsize$(-3.0)$ & 48.9 &\scriptsize$(+6.1)$ & 47.5 &\scriptsize$(+3.9)$ & 4.6  & 3.4  & 6.1  & 2.0 \\

\bottomrule
\end{tabular}

  \caption{
      Evaluation results of Qwen2-0.5B on CondaQA (top), and cue-frequency analysis for the best-performing further pre-trained model (bottom).
      Par., Sco., and Aff. denote paraphrase, scope edit, and affirmative edit.
      The lower section reports results on four CondaQA cue-frequency quartiles ($Q_1$--$Q_4$), where $Q_1$ contains the top 25\% most frequent negation cues. 
      $\Delta$ indicates the relative change with respect to the off-the-shelf model.
    }
  \label{t:condaqa_qwen_results}
\end{table*}
% %%%%%%%%%%%%%%%%%%%%%%%%%%%%%%%%%%%%%%%%%%%%%%%%%%%%%%%%%%%%%%
% Qwen Results
% %%%%%%%%%%%%%%%%%%%%%%%%%%%%%%%%%%%%%%%%%%%%%%%%%%%%%%%%%%%%%%
\section{Additional Results for Qwen}
\label{app:qwen_results}
\begin{table*}[t!]
  \centering
  \small
  \begin{tabular}{l r r@{ }l r@{ }l r@{ }l r@{ }l}
\toprule
           && \multicolumn{2}{c}{\textbf{NeQA}} &
              \multicolumn{2}{c}{\textbf{NevIR}} &
              \multicolumn{2}{c}{\textbf{NMoNLI}} &
              \multicolumn{2}{c}{\textbf{ScoNe-NLI}} \\ 
              \cmidrule(l){3-10}
              %\midrule
    
& \# Pars. & Acc. &($\Delta$\%) 
             & Acc. &($\Delta$\%) 
             & Acc. &($\Delta$\%) 
             & Acc. &($\Delta$\%) \\ 
             \midrule

Qwen2-0.5B   (off-the-shelf)     & 0.5B 
                     & 16.0 & 
                     & 49.3  & 
                     & 49.5 & 
                     & 50.0 & \\
~~~~~further pre-trained, NSPP with \\
~~~~~~~~~affixal cues    &
                     & \textbf{18.0} &\scriptsize$(+12.5)$ 
                     & \textbf{49.8} &\scriptsize$(+1.0)$ 
                     & 49.5 &\scriptsize$(+0.0)$ 
                     & \textbf{50.1} &\scriptsize$(+0.2)$ \\
~~~~~~~~~single-word cues&
                     & 17.0 &\scriptsize$(+6.3)$ 
                     & 49.6  &\scriptsize$(+0.6)$ 
                     & \textbf{50.0} &\scriptsize$(+1.0)$ 
                     & 49.7 &\scriptsize$(-0.8)$ \\
~~~~~~~~~multi-word cues &
                     & 16.0 &\scriptsize$(+0.0)$ 
                     & 49.1 &\scriptsize$(-0.4)$  
                     & 49.5 &\scriptsize$(+0.0)$ 
                     & \textbf{50.1} &\scriptsize$(+0.2)$ \\
~~~~~~~~~All cues    &
                     & 17.0 &\scriptsize$(+6.3)$ 
                     & 49.4  &\scriptsize$(+0.2)$  
                     & 49.5 &\scriptsize$(+0.0)$ 
                     & 50.0 &{\scriptsize $(+0.0)$} \\

\bottomrule
\end{tabular}
  \caption{
      Qwen results on four NLU Benchmarks.
      $\Delta$ indicates improvements with respect to the off-the-shelf model.
    }
  \label{t:qwen_4benchmarks_results}
\end{table*}

In addition to the Llama3.2-1B and Llama3.2-3B results presented in Tables~\ref{t:condaqa_main_results},~\ref{t:condaqa_freq_results}, and~\ref{t:other4benchmarks_results}, 
we also evaluate the smaller Qwen2-0.5B model from the Qwen family. 
We next analyze and discuss its results. 
Table~\ref{t:condaqa_qwen_results} presents the Qwen model's performance on the CondaQA benchmark. 
Notably, we combine the evaluations of accuracy, group consistency, and cue-frequency analysis into a single table. 
Table~\ref{t:qwen_4benchmarks_results} further reports its results on the other four natural language understanding (NLU) benchmarks.
Overall, despite its relatively small scale, Qwen2-0.5B still benefits from further pre-training on NegCue, with at least some NegCue variants consistently improving performance across benchmarks.

From Table~\ref{t:condaqa_qwen_results}, we observe that the Qwen model benefits most from further pre-training on the two less commonly studied negation types, namely multi-word and affixal negation. 
Among them, multi-word negation yields the largest improvement, slightly outperforming affixal negation by approximately 0.1\%. 
In contrast, consistent with our earlier observations on both Llama3.2-1B and Llama3.2-3B, single-word negation provides the smallest gains and even introduces slight negative effects in some settings. 
A similar pattern is also observed for the \textit{All Cues} corpus.
We speculate that this behavior is related to the relatively small scale of Qwen2-0.5B, which may make the model more sensitive to additional pre-training and hyperparameter choices. 
Since we do not perform extensive hyperparameter tuning, some settings may be suboptimal for smaller models.

Importantly, despite belonging to a different model family, Qwen exhibits a pattern consistent with the Llama models: 
performance remains strongest on the single-word subset across all negation types, supporting our earlier claim that current LLMs already possess stronger baseline capabilities for single-word negation understanding.
Moreover, the group consistency results show that all further pre-trained variants outperform the off-the-shelf model (All: 3.8\%), suggesting that NSPP pre-training improves robustness to negation-related edits. 
Finally, the cue-frequency analysis in the lower part of Table~\ref{t:condaqa_qwen_results} is also consistent with our previous observations: 
the largest improvements are achieved on the moderately frequent $Q_3$ cues ($\Delta$: +6.1\%).

The results of Qwen2-0.5B on the other four NLU benchmarks are reported in Table~\ref{t:qwen_4benchmarks_results}, where the best results are highlighted in bold. 
We again observe the previously discussed trend that models further pre-trained on the affixal negation corpus generally achieve stronger performance, including a substantial improvement of +12.5\% on NeQA.
In addition, as discussed earlier, the two NLI benchmarks remain considerably more challenging, even for larger LLMs. 
Our results show that Qwen2-0.5B achieves only limited improvements on these tasks; nevertheless, at least one NegCue variant consistently yields gains over the off-the-shelf model.

\textbf{Overall, NegCue pre-training remains beneficial even for smaller LLMs.} 
Most experimental findings on Qwen2-0.5B align with our earlier observations, further suggesting that negation-focused pre-training can improve negation understanding in language models.

% %%%%%%%%%%%%%%%%%%%%%%%%%%%%%%%%%%%%%%%%%%%%%%%%%%%%%%%%%%%%%%
% NSPP
% %%%%%%%%%%%%%%%%%%%%%%%%%%%%%%%%%%%%%%%%%%%%%%%%%%%%%%%%%%%%%%
\section{NSPP Further Pre-training Details}
\label{app:nspp_experimental_details}

\begin{figure*}[t]
\small

\begin{tcolorbox}[
    colback=gray!5,
    colframe=gray!70,
    boxrule=0.5pt,
    arc=2pt,
    left=6pt,
    right=6pt,
    top=4pt,
    bottom=4pt
]

\textbf{Instruction:}

\vspace{0.3em}

You are a helpful assistant. In this task, you will be given the first sentence ($S_1$) from a pair of consecutive sentences.

\vspace{0.3em}

Your goal is to determine whether the second sentence ($S_2$) would contain a negation with respect to the first sentence ($S_1$).

\vspace{0.3em}

The answer should be either ``yes'' or ``no''.

\vspace{0.3em}

Only output one of the following labels: [YES] or [NO].

\vspace{0.6em}

\noindent\rule{\textwidth}{0.4pt}

\vspace{0.6em}

\textbf{Input:}

\vspace{0.3em}

\textbf{Sentence 1:} This started life as a modest hall house on a site overlooking the Dour Burn in the 13th century.

\vspace{0.6em}

\noindent\rule{\textwidth}{0.4pt}

\vspace{0.6em}

\textbf{Response:}

\vspace{0.3em}

[NO]

\end{tcolorbox}

\caption{
Zero-shot prompt used to further pre-train LLMs on the NSPP task with our NegCue corpus.
}

\label{fig:nspp_prompt}

\end{figure*}
In this section, we describe additional details of our NSPP further pre-training setup and experimental configuration. 
We focus on two encoder-only LMs, BERT-large~\cite{devlin-etal-2019-bert} and RoBERTa-large~\cite{liu2019roberta}, as well as several decoder-only LLMs across multiple scales, ranging from 0.5B to 3B parameters, including Qwen2-0.5B~\cite{yang2024qwen2technicalreport}, Llama3.2-1B, and Llama3.2-3B~\cite{grattafiori2024llama}. All models are trained on a single H100 GPU with 80GB of memory.

For the encoder-only Transformer models, namely BERT-large and RoBERTa-large, 
we perform self-supervised training on the NSPP objective for two epochs, with early stopping enabled based on validation performance.
Both models are further pre-trained with the same hyperparameter settings, using a batch size of 512, a learning rate of 1e-6, 
and a linear learning rate scheduler with warmup, following standard practice in Transformer-based models~\cite{vaswani2017attention}.

% In addition to these two models, we also pre-trained and evaluated the relatively smaller BERT-base and RoBERTa-base models. 
% After further pre-training on our corpus, both models show substantial improvements on CondaQA compared to their off-the-shelf counterparts.

\begin{figure*}[t]
    \centering
    \includegraphics[width=\textwidth]{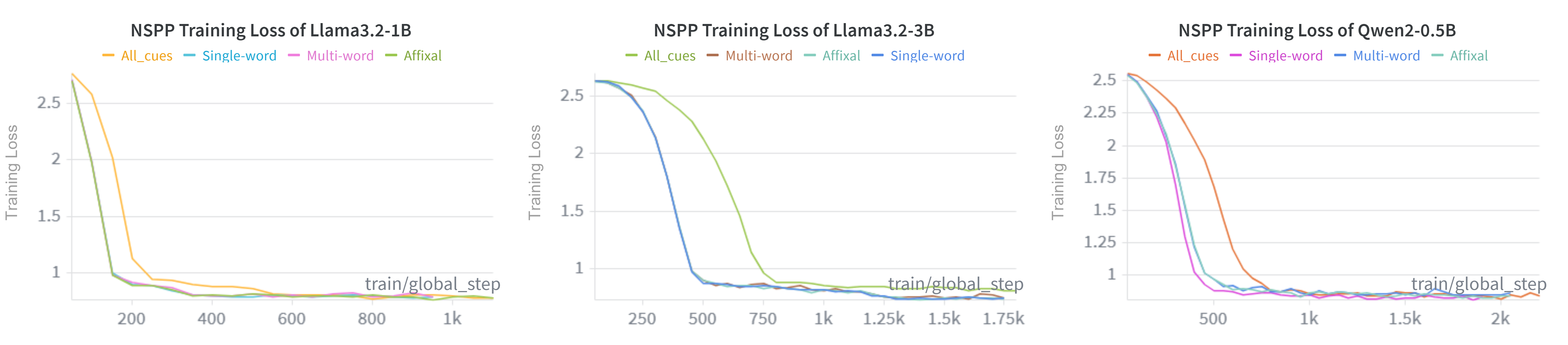}
    \caption{NSPP training loss curves of the three LLMs across training steps.}
    \label{fig:nspp_training_loss}
\end{figure*}

For smaller large language models (LLMs), we adopt a zero-shot prompt-based formulation of the NSPP task and 
further pre-train the models using QLoRA~\cite{dettmers2023qlora} with 4-bit quantization to reduce computational and memory costs 
while enabling efficient training on limited hardware resources.
The zero-shot prompt used for NSPP pre-training is shown in Figure~\ref{fig:nspp_prompt}.
All LLMs are fine-tuned with a batch size of 16 using a cosine learning rate scheduler and a LoRA rank of 32. 
We use a learning rate of 2e-4 for Llama3.2-1B and 1e-5 for both Qwen2-0.5B and Llama3.2-3B, following widely adopted default settings.

For fair comparison, we use the same hyperparameter settings when training a given model on the four NegCue sub-corpora.
Training is conducted for a single epoch, as these models converge rapidly and reach a performance plateau early, 
with only marginal decreases in validation loss thereafter. 
To mitigate overfitting, we apply standard early stopping with a patience of 3, and additionally introduce a convergence-based stopping criterion, 
terminating training when the validation loss decreases by less than 0.01 for three consecutive evaluations.
Under these settings, training time ranges from approximately 3 to 16 hours, highlighting that further pre-training even relatively small LLMs still incurs non-trivial computational and time costs.

Figure~\ref{fig:nspp_training_loss} presents the training loss curves of the three LLMs across training steps, with different colors corresponding to different training corpora. 
Across all settings, the training loss consistently decreases and gradually converges as training progresses, suggesting that the NSPP objective contains stable and learnable training signals.

% %%%%%%%%%%%%%%%%%%%%%%%%%%%%%%%%%%%%%%%%%%%%%%%%%%%%%%%%%%%%%%
% Benchmarks
% %%%%%%%%%%%%%%%%%%%%%%%%%%%%%%%%%%%%%%%%%%%%%%%%%%%%%%%%%%%%%%
\section{Benchmark Details}
\label{app:benchmarks}
Next, we provide a detailed description of the benchmarks used for evaluation, and Table~\ref{t:benchmarks_samples} summarizes the dataset statistics for each benchmark.

\textbf{CondaQA}
~\cite{ravichander-etal-2022-condaqa} is a question answering benchmark designed to check
the ability to reason about negation in context.
Each sample consists of a question paired with a passage containing a negated statement relevant to the question.
Answers take one of several forms: \textit{Yes}, \textit{No}, \textit{Don’t know}, or a span from the passage. 
Each question is paired with the original passage (with negation) from Wikipedia
and three modified versions:
a (meaning-preserving) paraphrase,
a version modifying the scope of the negation,
and 
a version removing the negation.
Note that the same question often has different answers depending on which passage is given in the input.
In addition to \textit{accuracy},
the authors of CondaQA propose as an evaluation metric \textit{group consistency}:~%
whether the answers from all (or a subset) of passages (original and modified) are correct.
It contains 7,240 samples in the test split,
and 5,832 and 1,110 in the train and validation splits.
%samples are organized into contrastive sets consisting of an original passage and three minimally edited variants, corresponding to paraphrase (\textit{Par.}), scope change (\textit{Sco.}), and affirmative edits (\textit{Aff.}). 
%Group consistency measures whether a model answers a question correctly for all passages within the same contrastive set.

\textbf{NeQA}
~\cite{zhang-etal-2023-beyond} is a multiple-choice commonsense question answering benchmark that involves negation. 
Each sample consists of a negated question paired with two answer choices: one correct and one incorrect. 
NeQA exhibits inverse scaling behavior~\cite{wei-etal-2023-inverse, mckenzie2024inversescalingbiggerisnt},
meaning that performance does not consistently improve as model size increases. 
NeQA was specifically designed to evaluate negation understanding in a simple setting and shows that current LLMs still struggle with negation.

\textbf{NevIR}
~\cite{weller-etal-2024-nevir} is a benchmark that evaluates negation understanding in information retrieval,
specifically, ranking documents given a query.
It contains contrastive query–document pairs where the only difference in the queries is whether they contain negation
(e.g., What should I do / avoid if my friend has a seizure?).
Document rankings change depending on whether the query contains negation.
The original task evaluates retrieval models by ranking documents according to similarity scores, 
and has not been evaluated on encoder-only or decoder-only language models.

\textbf{NMoNLI and ScoNe-NLI}
are negation-focused lexical entailment benchmarks for natural language inference (NLI). 
Both are binary classification tasks and consider two labels: entailment and neutral. %~(no contradictions, unlike other NLI benchmarks).
NMoNLI extends SNLI~\cite{bowman-etal-2015-large} by introducing negation into each sample~\cite{geiger-etal-2020-neural}.
The negation is restricted to scope over the substituted token span,
thus it affects the entailment relation.
ScoNe-NLI~\cite{she-etal-2023-scone} further extends it by introducing contrast sets that systematically vary the scope and interaction of negation.
Specifically, ScoNe-NLI consists of six contrast splits, each containing 1,002 training samples and 200 test samples, designed to isolate the effects of negation scope on entailment relations. The six splits correspond to the following cases:
\begin{itemize}[noitemsep, topsep=0pt]
    \item \textit{no\_negation}: no negation is present.
    \item \textit{one\_not\_scoped}: one negation that does not scope over the relevant lexical item.
    \item \textit{two\_not\_scoped}: two negations, neither of which scopes over the relevant lexical item.
    \item \textit{two\_scoped}: two negations, where one scopes over the relevant lexical item, but the second scopes over the first, canceling it out.
    \item \textit{one\_scoped}: one negation that scopes over the relevant lexical item.
    \item \textit{one\_scoped\_one\_not\_scoped}: two negations, but only one affects the relevant lexical item.
\end{itemize}
Together, these six splits yield a total of 6,012 training samples and 1,200 test samples across the full benchmark. 
Table~\ref{t:finetune_on_scone} presents the results of our further pre-trained LMs on all six splits, and Table~\ref{t:other4benchmarks_results} reports the average accuracy.

Overall, these two benchmarks were designed to evaluate performance on NLI tasks involving complex negation in the premise or hypothesis.

\begin{table}[t!]
  \centering
  \small
  \begin{tabular}{lrrr}
  \toprule
            & \textbf{Training} & \textbf{Validation} & \textbf{Test} \\
  \midrule
  CondaQA   & 5,832    & 1,110       & 7,240 \\
  NeQA      & 374     & 54         & 100 \\
  NevIR     & 948     & 225        & 1,383 \\
  NMoNLI    & 1,002    & n/a        & 200 \\
  ScoNe-NLI & 6,012    & n/a        & 1,200 \\
  SNLI      & 366,180 & 6,564       & n/a \\
  \bottomrule
  \end{tabular}
  \caption{   
      The number of samples in the training, validation, and test sets of the benchmarks.
      }
  \label{t:benchmarks_samples}
\end{table}

% %%%%%%%%%%%%%%%%%%%%%%%%%%%%%%%%%%%%%%%%%%%%%%%%%%%%%%%%%%%%%%
% Evaluations
% %%%%%%%%%%%%%%%%%%%%%%%%%%%%%%%%%%%%%%%%%%%%%%%%%%%%%%%%%%%%%%
\section{Evaluation on Downstream Tasks}
\label{app:evaluation_details}
In this section, we describe the evaluation setup for both encoder-only LMs and LLMs on the aforementioned benchmarks, 
including how we fine-tune LMs on downstream tasks and the prompt templates used for zero-shot evaluation with LLMs. 
For all LLM evaluations, we set the temperature to zero and use deterministic decoding (\texttt{do\_sample=False} without top-$p$ sampling), yielding almost identical outputs across repeated runs. Table~\ref{t:benchmarks_hyperparameters} reports the hyperparameter settings used for fine-tuning LMs across benchmarks.

\begin{table}[t!]
  \centering
  \small
  \begin{tabular}{lrrr}
  \toprule
            & \textbf{Learning Rate} & \textbf{Batch Size} & \textbf{Epochs} \\
  \midrule
% CondaQA   & \{5e-6, 1e-5\} & \{8, 16\} & 10 \\
CondaQA     & 5e-6, 1e-5 & 8, 16 & 10 \\
  NeQA      & 1e-5 & 16 & 8 \\
  NevIR     & 2e-5 & 16 & 20 \\
  SNLI      & 2e-5 & 16 & 1 \\
  % ScoNe-NLI & 2e-5 & 16 & 1 \\
  \bottomrule
  \end{tabular}
  \caption{   
      Hyperparameters used for fully supervised fine-tuning of LMs across benchmarks.
      For CondaQA, we perform grid search to select the optimal hyperparameter configuration. 
      For NMoNLI and ScoNe-NLI, models are instead fine-tuned on SNLI, as in-domain fine-tuning leads to near-saturated performance.
      }
  \label{t:benchmarks_hyperparameters}
\end{table}

%%%%%%%%%%%% condaqa
\subsection{Evaluation on CondaQA}
\label{app:condaqa_details}
CondaQA provides training, validation, and test splits, with dataset statistics summarized in Table~\ref{t:benchmarks_samples}. 
In addition to the evaluation metrics reported in the paper, namely accuracy and group consistency, 
we further evaluate model performance across different negation types and cue frequency ranges. 
Specifically, we partition the CondaQA test set into three subsets based on cue type: affixal, single-word, and multi-word. 
We also group test samples into four quartiles ($Q_1$–$Q_4$) according to the frequency of the negation cues they contain, ordered from most frequent to least frequent.

For encoder-only LMs further pre-trained on our corpus, including BERT-large and RoBERTa-large, we fine-tune the models on the CondaQA training set. 
We perform grid search over learning rates \{5e-6, 1e-5\} and batch sizes \{8, 16\}, selecting the best configuration for each model based on validation performance. 
Following the original CondaQA setting, we fine-tune the models for 10 epochs and evaluate them on the CondaQA test set using the metrics described above.

In contrast, for LLMs, we use the five test splits provided by the original authors for in-context learning (ICL) evaluation. 
Specifically, we evaluate two scales of Llama and Qwen models under a zero-shot setting, using the same prompt as in the original paper (shown in Figure~\ref{fig:condaqa_prompt}), with the task description following~\citet{wang2022benchmarking}.

% Specifically, we evaluate LLaMA3.2-1B and Qwen2-0.5B under a zero-shot setting, using the same prompt as in the original paper (shown in Table~\ref{t:condaqa_prompt}), with the task description following~\citet{wang2022benchmarking}.

%%%%%%%%%%%%%%%% neqa
\subsection{Evaluation on NeQA}
\label{app:neqa_details}
NeQA is a dataset consisting of 1,718 commonsense multiple-choice questions with negation sourced from several existing benchmarks. 
Unlike CondaQA, NeQA does not provide predefined training, validation, or test splits. 
In the original work, the authors randomly sampled 100 examples from the dataset to construct a test set for evaluating LLMs. 
Encoder-only language models were not evaluated.
To ensure consistency and comparability with prior work, we adopt the same 100 samples as our test set. 
For encoder-only LMs, we fine-tune the models directly on NeQA rather than on external QA datasets to enable question-answering capabilities.
Specifically, we construct training and validation splits from the remaining NeQA samples based on the dataset metadata, with dataset statistics reported in Table~\ref{t:benchmarks_samples}.
To prevent data leakage, we ensure that no test sample appears in either the training or validation set.

% \noindent
% \textbf{Training and Validation Split Construction}
Each NeQA sample consists of a question paired with two answer options. 
To adapt the task for encoder-only LMs further pre-trained on the NSPP, we convert each original sample into two binary classification examples. 
Each converted example contains a declarative statement formed by combining the question with one answer option, along with a label indicating whether the option is correct.
For example, given the question \textit{“Birds cannot ?”} with options \textit{(A) fly} and \textit{(B) ulster}, 
where the correct answer is \textit{(B)}, we convert the sample into two examples: 
\textit{“Birds cannot fly.”} labeled as \textit{incorrect}, and \textit{“Birds cannot ulster.”} labeled as \textit{correct}.
As a result, the transformed dataset contains twice as many training samples as the original NeQA samples. 
As shown in Table~\ref{t:benchmarks_samples}, we use 374 NeQA samples to construct the training data, yielding a total of 748 binary samples for LM fine-tuning.

% \textbf{Evaluation Settings.}
After building the training and validation sets, we fine-tune BERT-large and RoBERTa-large using a unified set of hyperparameters. 
Specifically, we use a learning rate of 1e-5, a batch size of 16, and fine-tune the models for 8 epochs. 
The fine-tuned models are then evaluated on the same 100 test samples used in the original NeQA work, and accuracy is reported for comparison.
In contrast, LLMs are evaluated under a zero-shot setting. 
To mitigate prompt sensitivity in language model evaluation~\cite{burns2022discovering}, we adopt the same prompt used in the original NeQA work, as shown in Figure~\ref{fig:llm_prompts}. 
This ensures a fair and consistent comparison with prior results.

%%%%%%%%%%%%%%%%%%%%% NevIR
\subsection{Evaluation on NevIR}
\label{app:nevir_details}

The NevIR benchmark was originally designed for information retrieval, where each sample consists of two queries and two documents. 
Since NSPP is formulated as a binary classification task, language models further pre-trained on NSPP cannot be directly evaluated on the original benchmark format. 
To address this, we first decompose each original sample into two triplet instances, each containing one query and two candidate documents. 
We then use the Sentence-Transformers framework to fine-tune the further pre-trained models on the NevIR training set.
Following the original NevIR work, where several information retrieval models are fine-tuned with a learning rate of 2e-5 for 20 epochs, 
we adopt the same hyperparameter settings to fine-tune our encoder-only LMs, using a batch size of 16, and evaluate the models using accuracy.
For LLM evaluation, we convert NevIR samples into instruction-style prompts. The prompt used in our experiments is shown in Figure~\ref{fig:nevir_prompt}.

%%%%%%%%%%%%%%%%%%%%%%% NMoNLi& ScoNe-NLI
\subsection{Evaluation on NLI Benchmarks}
\label{app:nli_details}

\begin{table*}[t]
\centering
\small
\begin{tabular}{l cccccc}
\toprule
 % & No\_negation & One\_not\_scoped & Two\_not\_scoped & Two scoped & One\_scoped & One scoped one not \\ %& Avg. Acc. \\
  & \textbf{No}       & \textbf{One}        & \textbf{Two}        & \textbf{Two}    & \textbf{One}    & \textbf{One Scoped}     \\ %& Avg. Acc. \\
  & \textbf{Negation} & \textbf{Not Scoped} & \textbf{Not Scoped} & \textbf{Scoped} & \textbf{Scoped} & \textbf{One Not Scoped} \\ %& Avg. Acc. \\
\midrule

BERT-large (off-the-shelf )
& \hphantom{0}93.5 & \hphantom{0}97.0 & \hphantom{0}94.0 & \hphantom{0}97.5 & \hphantom{0}98.5 & \hphantom{0}98.5 \\ % & 96.5 \\
~~~+ NSPP, Affixal
& \hphantom{0}99.0 & \hphantom{0}99.0 & \hphantom{0}98.0 & \hphantom{0}99.5 & \hphantom{0}99.0 & 100.0 \\ % & 99.1 \\
~~~+ NSPP, Single-word
& \hphantom{0}98.5 & \hphantom{0}99.5 & \hphantom{0}99.5 & 100.0 & \hphantom{0}99.0 & \hphantom{0}99.0 \\ % & 99.3 \\
~~~+ NSPP, Multi-word
& \hphantom{0}99.0 & \hphantom{0}99.0 & \hphantom{0}99.0 & 100.0 & \hphantom{0}99.5 & \hphantom{0}99.0 \\ % & 99.3 \\
~~~+ NSPP, All cues
& \hphantom{0}99.0 & \hphantom{0}99.5 & \hphantom{0}98.0 & \hphantom{0}99.5 & \hphantom{0}99.0 & 100.0 \\ % & 99.2 \\

\midrule
RoBERTa-large (off-the-shelf)
& 100.0 & 100.0 & \hphantom{0}97.5 & \hphantom{0}99.5 & \hphantom{0}99.5 & \hphantom{0}99.5 \\ % & 99.3 \\
~~~+ NSPP, Affixal
& 100.0 & 100.0 & 100.0 & 100.0 & 100.0 & 100.0 \\ % & 100.0 \\
~~~+ NSPP, Single-word
& 100.0 & 100.0 & 100.0 & 100.0 & 100.0 & 100.0 \\ % & 100.0 \\
~~~+ NSPP, Multi-word
& 100.0 & 100.0 & 100.0 & 100.0 & 100.0 & 100.0 \\ % & 100.0 \\
~~~+ NSPP, All cues
& 100.0 & 100.0 & 100.0 & 100.0 & 100.0 & 100.0 \\ % & 100.0 \\

\bottomrule
\end{tabular}
\caption{
    Accuracy (\%) on the six ScoNe-NLI test splits for BERT-large and RoBERTa-large fine-tuned on the ScoNe-NLI training set. 
    All models, including off-the-shelf baselines, achieve near-perfect results across all splits.}
\label{t:finetune_on_scone}
\end{table*}

We evaluate two NLI benchmarks, NMoNLI and ScoNe-NLI, both of which provide training and test splits. 
The corresponding dataset statistics are reported in Table~\ref{t:benchmarks_samples}. 
Unlike NeQA and NevIR, both NMoNLI and ScoNe-NLI are already formulated as binary classification tasks and therefore require no task reformulation. 
We directly fine-tune BERT-large and RoBERTa-large on the training sets of these two benchmarks and evaluate the resulting models on their respective test sets.

However, we observe that after fine-tuning on the in-domain training data, LMs achieve near-saturated performance on the test sets, leaving little room for meaningful comparison with prior baselines. 
For example, RoBERTa-large attains 100\% accuracy on all six ScoNe-NLI splits after fine-tuning. 
The detailed results for BERT-large and RoBERTa-large fine-tuned on ScoNe-NLI are reported in Table~\ref{t:finetune_on_scone}.
To obtain more informative comparisons, we instead fine-tune both LMs on the widely used SNLI dataset~\cite{bowman-etal-2015-large}. 
SNLI is a widely used and influential NLI benchmark that shares the same task objective as our evaluation datasets. 
Moreover, both ScoNe-NLI and NMoNLI are derived from SNLI. 
We therefore fine-tune the models on this more general and closely related dataset, 
allowing us to better isolate the effects of our NSPP pre-training across different negation types in downstream NLI tasks.

The models are fine-tuned for one epoch with a learning rate of 2e-5 and a batch size of 16, and are then directly evaluated on the NMoNLI and ScoNe-NLI test sets. 
For LLM evaluation, we convert the training samples into instruction-style prompts and perform zero-shot evaluation on the test sets. 
The prompts used for NMoNLI and ScoNe-NLI are shown in Figure~\ref{fig:llm_prompts}.
We report accuracy on NMoNLI, and for ScoNe-NLI, we report the average accuracy across all six splits.

\begin{table*}[t]
  \centering
  \small
  \begin{tabular}{lcccc}
\toprule
\textbf{Model} & \textbf{Entailment} & \textbf{Not Entailment} & \textbf{All} & \textbf{Strict Accuracy} \\
\midrule
RoBERTa-MNLI        & 0.648 & 0.684 & 0.670 & 0.250 \\
RoBERTa-NegCue-MNLI & \textbf{0.710} & \textbf{0.708} & \textbf{0.709} & \textbf{0.271} \\
BERT-NegCue-MNLI    & 0.639 & 0.607 & 0.623 & 0.104 \\
\bottomrule
\end{tabular}
  \caption{
        Results on NaN-NLI. We report F1 scores for Entailment, Not Entailment, and All under the Binary setting, together with Strict Accuracy. Best results are shown in bold.
  }
  \label{t:nan_nli_results}
\end{table*}

% %%%%%%%%%%%%%%%%%%%%%%%%%%%%%%%%%%%%%%%%%%%%%%%%%%%%%%%%%%%%%%
% NaN-NLI Results
% %%%%%%%%%%%%%%%%%%%%%%%%%%%%%%%%%%%%%%%%%%%%%%%%%%%%%%%%%%%%%%
\section{Additional Evaluation on NaN-NLI}
\label{app:nan_nli_results}

In addition to the five benchmarks discussed in the main paper, we further evaluate our encoder-only models on NaN-NLI~\cite{truong-etal-2022-another}, 
a linguistically grounded NLI test suite designed to assess sub-clausal negation through minimally different premise–hypothesis pairs.
Specifically, we use BERT-large and RoBERTa-large models first continued-pretrained with NSPP on the full NegCue corpus and then fine-tuned on the MNLI dataset~\cite{williams-etal-2018-broad} before evaluation on NaN-NLI. 
We compare against the RoBERTa-MNLI result reported in the original NaN-NLI paper. 
Since the original study evaluates only encoder-based models, we restrict this additional evaluation to encoder-only models for direct comparison. 
Following the original benchmark, we report results under the Binary and Strict settings. 
In the Binary setting, Contradiction and Neutral are merged into Not Entailment, with F1 scores reported for Entailment, Not Entailment, and All. 
Strict accuracy considers a premise correct only if all of its associated hypotheses are assigned the correct labels.
As shown in Table~\ref{t:nan_nli_results}, our RoBERTa-NegCue-MNLI model outperforms the reported RoBERTa-MNLI baseline across all Binary metrics, increasing the overall F1 score from 0.670 to 0.709. 
Strict accuracy also improves from 0.250 to 0.271. 
These results provide further evidence that our pre-training on NegCue improves negation understanding.

\begin{figure*}[t]
\small
\begin{tcolorbox}[
    colback=gray!5,
    colframe=gray!70,
    boxrule=0.5pt,
    arc=2pt,
    left=6pt,
    right=6pt,
    top=4pt,
    bottom=4pt
]

In this task, you're expected to write answers to questions involving reasoning about negation. The answer to the question should be ``yes'', ``no'', ``don't know'' or a phrase in the passage. Questions can have only one correct answer.

\vspace{0.5em}

\textbf{Passage:} In October 2010, Daniela Soleri - Paolo Soleri's daughter - resigned from the Cosanti Foundation board, citing abuse by her father. She stated that some of Soleri's inner circle had been told decades earlier, but they remained actionless about it at the time. After the resignation, Soleri stepped down as chairman, but the board made no public statement on the reasons.

\vspace{0.5em}

\textbf{Question:} Is it likely that people in Paolo's inner circle talked to him about the abuse of his daughter?

\vspace{0.5em}

\textbf{Answer:}

\end{tcolorbox}

\caption{Zero-shot prompt example used to evaluate LLMs with CondaQA.}
\label{fig:condaqa_prompt}
\end{figure*}
\begin{figure*}[t]
\small
\begin{tcolorbox}[
    colback=gray!5,
    colframe=gray!70,
    boxrule=0.5pt,
    arc=2pt,
    left=6pt,
    right=6pt,
    top=4pt,
    bottom=4pt
]

In this task, you will be given two candidate documents and a query. Your goal is to decide which document answers the query.

\vspace{0.5em}

\textbf{\#\#DOCUMENT 1:}

In 2012 Dyson published (with William H. Press) a fundamental new result about the prisoner's dilemma in the Proceedings of the National Academy of Sciences of the United States of America. He wrote a foreword to a treatise on psychic phenomena in which he concluded that ``ESP is real... but cannot be tested with the clumsy tools of science''.

\vspace{0.5em}

\textbf{\#\#DOCUMENT 2:}

In 2012 Dyson published (with William H. Press) a fundamental new result about the prisoner's dilemma in the Proceedings of the National Academy of Sciences of the United States of America. He wrote a foreword to a treatise on psychic phenomena in which he concluded that ``ESP is real... and has been tested with the clumsy tools of science''.

\vspace{0.5em}

Output ``1'' if the first document contains the answer, or ``2'' if the second document contains the answer.

\vspace{0.5em}

You MUST answer with exactly one character: ``1'' or ``2''.

\vspace{0.5em}

\textbf{Query:}

Who said that ESP is real though not testable by science? Only output ``1'' or ``2''.

\vspace{0.5em}

\textbf{Response:}

\end{tcolorbox}

\caption{Zero-shot prompt example used to evaluate LLMs on NevIR.}
\label{fig:nevir_prompt}

\end{figure*}
\begin{figure*}[t]
\small
\begin{tcolorbox}[
    colback=gray!5,
    colframe=gray!70,
    boxrule=0.5pt,
    arc=2pt,
    left=6pt,
    right=6pt,
    top=4pt,
    bottom=4pt
]

\textbf{NeQA}

\vspace{0.3em}

The following are multiple choice questions (with answers) about common sense.

\vspace{0.3em}

\textbf{Question:} Crying is not part of ? \\
A. life \\
B. love \\
\textbf{Answer:}

\vspace{0.6em}

\hrule

\vspace{0.6em}

\textbf{ScoNe-NLI}

\vspace{0.3em}

\textbf{Question:} Assume that the man does not own a dog and does not own a cat. Is it then definitely true that the man does not own an animal and does not own a cat? Only answer yes or no.

\vspace{0.3em}

\textbf{Answer:}

\vspace{0.6em}

\hrule

\vspace{0.6em}

\textbf{NMoNLI}

\vspace{0.3em}

This is a logical and commonsense reasoning exam. The answer must be Yes or No. Only output Yes or No. 

\vspace{0.3em}

\textbf{Question:} Assume that the man does not own a dog. Is it definitely true that the man does not own a mammal?

\vspace{0.3em}

\textbf{Answer:}

\end{tcolorbox}

\caption{Zero-shot prompt examples used for evaluation on NeQA, ScoNe-NLI, and NMoNLI.}
\label{fig:llm_prompts}

\end{figure*}
\end{document}